\documentclass{article}

\PassOptionsToPackage{numbers, compress}{natbib}
\usepackage[preprint]{neurips}

\usepackage{textcomp}

\usepackage{blindtext}
\usepackage{CJKutf8}
\usepackage{titlesec}
\usepackage{setspace}
\usepackage{algorithm}
\usepackage{algorithmic}
\usepackage{subcaption}

\usepackage[utf8]{inputenc} 
\usepackage[T1]{fontenc} 
\usepackage{hyperref} 
\usepackage{url} 
\usepackage{booktabs} 
\usepackage{amsfonts} 
\usepackage{nicefrac} 
\usepackage{microtype} 
\usepackage{xcolor} 
\usepackage{graphicx}
\usepackage{amssymb}
\usepackage{mathtools}
\usepackage{amsthm}
\usepackage{multirow}
\usepackage{colortbl}
\usepackage{enumitem}
\usepackage{tocloft}

\definecolor{softblue}{RGB}{173, 216, 230} 
\definecolor{lightgreen}{RGB}{144, 238, 144} 
\definecolor{peach}{RGB}{255, 218, 185} 
\definecolor{lavender}{RGB}{230, 230, 250} 

\newcommand{\CC}{\mathcal{C}}
\newcommand{\DD}{\mathcal{D}}
\newcommand{\XX}{\mathbf{X}}
\newcommand{\zz}{\mathbf{z}}
\newcommand{\Real}{\mathbb{R}}

\newcommand{\mypar}[1]{\noindent\textbf{#1}}
\newcommand{\method}{\emph{DART}\textsuperscript{3}}

\title{\method: Leveraging Distance for Test Time Adaptation in Person Re-Identification}

\author{%
 Rajarshi Bhattacharya $^{1}$ \\
 \texttt{rajarshi.bhattacharya.1@ens.etsmtl.ca}\\
 \And
 Shakeeb Murtaza $^{1}$ \\
 \texttt{shakeeb.murtaza.1@ens.etsmtl.ca} \\
 \And
 Christian Desrosiers $^{1}$ \\
 \texttt{christian.desrosiers@etsmtl.ca} \\
 \And
 Jose Dolz $^{1}$ \\
 \texttt{jose.dolz@etsmtl.ca} \\
 \And
 Maguelonne Heritier $^{2}$ \\
 \texttt{mheritier@genetec.com} \\
 \And
 Eric Granger $^{1}$ \\
 \texttt{eric.granger@etsmtl.ca} \\
 \AND \\
 $^{1}$ LIVIA, École de technologie supérieure \\
 $^{2}$ Genetec Inc.
}

\makeatletter

\begin{document}

\maketitle

\begin{abstract}
Person re-identification (ReID) models are known to suffer from camera bias, where learned representations cluster according to camera viewpoints rather than identity, leading to significant performance degradation under (inter-camera) domain shifts in real-world surveillance systems when new cameras are added to camera networks. State-of-the-art test-time adaptation (TTA) methods, largely designed for classification tasks, rely on classification entropy-based objectives that fail to generalize well to ReID, thus making them unsuitable for tackling camera bias. In this paper, we introduce \method, a TTA framework specifically designed to mitigate camera-induced domain shifts in person ReID. \method (\textbf{D}istance-\textbf{A}ware \textbf{R}etrieval \textbf{T}uning at \textbf{T}est \textbf{T}ime) leverages a distance-based objective that aligns better with image retrieval tasks like ReID by exploiting the correlation between nearest-neighbor distance and prediction error. Unlike prior ReID-specific domain adaptation methods, \method requires no source data, architectural modifications, or retraining, and can be deployed in both fully black-box and hybrid settings. Empirical evaluations on multiple ReID benchmarks indicate that \method and \method LITE, a lightweight alternative to the approach, consistently outperforms state-of-the-art  TTA baselines, making for a viable option to online learning to mitigate the adverse effects of camera bias.

\end{abstract}


\section{Introduction}
\vspace{-8pt}
Person re-identification (ReID), the task of matching a query identity with a set of gallery candidates, has been widely adopted in real-time video surveillance systems around the world. Despite significant advances across various challenging scenarios, such as under occlusions \cite{somers2023body}, variations in clothing of the person \cite{jin2022cloth}, cross-modal settings \cite{wu2023unsupervised} and text guided settings \cite{li2023clip}, most methods fail to address specific biases inherent to these ReID models. Camera bias is a prevalent issue in person ReID, where query and gallery representations tend to cluster based on the camera the images were captured from, rather than 
person identity \cite{song2025exploring, cho2022part, lee2023camera, chen2021ice, wang2021camera, gu20201st, luo2020generalizing}. This issue hinders the scaling of existing surveillance systems through the addition of new cameras.

\begin{figure*}[!t]
 \centering
 \includegraphics[width=0.8\linewidth]{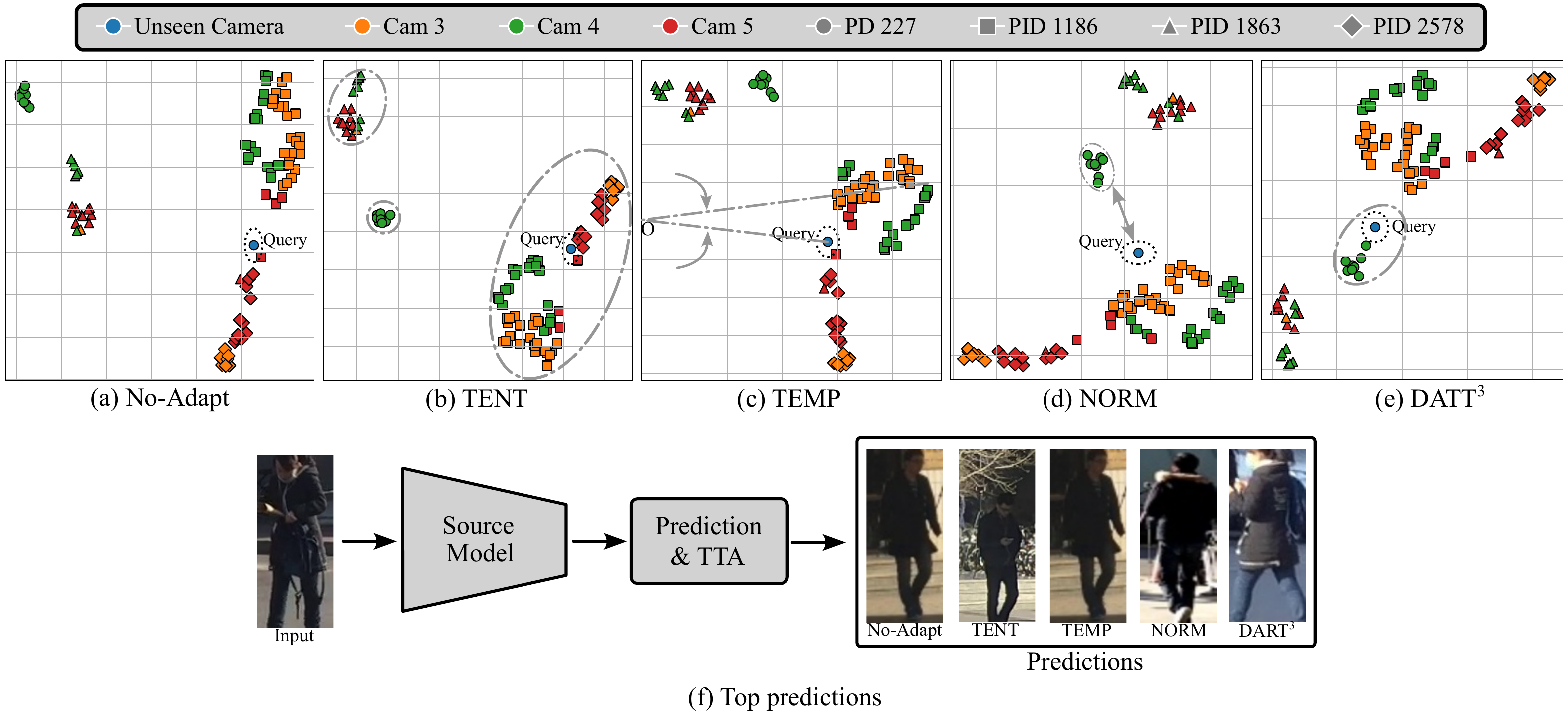}
 \caption{A failure case where the source model misidentifies a query image due to camera bias—favoring views with similar pose, lighting, and perspective. (a)–(e) show t-SNE visualizations of the top feature matches for different test-time adaptation methods. Unseen camera (a) refers to a query sample taken from a camera not used for training the source model. TENT \cite{wang2020tent} (b) over-condenses erroneous clusters; TEMP \cite{adachi2024test} (c) aligns features angularly but still misidentifies; Camera Normalization \cite{song2025exploring} (d) reduces bias towards background artifacts yet fails to recover identity. Our method (e) achieves correct retrieval despite view and lighting differences. (f) shows the top predictions for each adaptation method.}
 \label{fig:teaser}
 \vspace{-10pt}
\end{figure*}

Test-time adaptation (TTA) provides a compelling approach towards solving domain shift related problems, largely due of its inherent source-free and unsupervised nature. Although numerous TTA methods have been proposed to mitigate domain shifts in classification tasks \cite{wang2020tent, zhou2021bayesian, niu2022efficient}, their application to person ReID remains relatively underexplored, with only a few studies addressing this area \cite{adachi2024test, almansoori2023anchor, han2022generalizable}. However, these studies largely overlook the problem of domain shift caused by camera biases, as illustrated in \autoref{fig:teaser}. State-of-the-art  TTA approaches also tend to rely heavily on entropy-based adaptation objectives, mirroring strategies traditionally used in image classification.  As shown in this work, this approach is suboptimal, as the error rate does not consistently correlate with changes in prediction entropy.

To address these limitations, we introduce \method—a novel TTA method that tackles the challenge of camera bias in person ReID, requiring no modifications to the source model architecture or pretraining procedure.  \method is based on the observation that, as the distance between the query and the nearest gallery samples increases, the error rate of the source pre-trained model also rises, as shown later in \autoref{fig:error_rate_curve}. We formulate a mathematical model to describe camera bias, and define a post-hoc scale-shift operation based on this model. Our method adaptively updates only a small subset of external parameters ($\sim$30k-50k), with or without modifying batch normalization weights, thereby opening the possibility of preserving the source model as a \textit{black-box} in applications where model privacy is crucial.


Our contributions can be summarized as follows:
\vspace{-3pt}


\begin{itemize}[leftmargin=2em,itemsep=0.2em]
    \item We propose \method, a novel TTA method person ReID addressing domain shifts caused by camera bias, which uses a distance-based TTA objective, specialized for ReID tasks. Our experimental observations show that top-k Euclidean distance is better suited as a TTA objective than entropy.
    \item A mathematical model for the camera bias is formalized, which further motivates our choice of external scale-shift adaptation parameters.
    \item Extensive empirical evaluations on several challenging datasets validate our hypotheses about the training objective and the nature of camera bias.  Our proposed \method can improve performance over state-of-the-art TTA methods, also be implemented in a perfectly \textit{black-box} setting, costing only a fraction of the inference time.
\end{itemize}

\section{Related Works}
\vspace{-8pt}

\mypar{Camera Bias.} 
Camera-aware person ReID methods \cite{luo2020generalizing, zhuang2020rethinking, zhang2021unsupervised, wang2021camera, chen2021ice, cho2022part, lee2023camera} leverage camera information during training to learn more robust features for person re-identification at test time. However, these methods rely on specialized pipelines and training strategies that require supervised training on the target data. The approaches in \cite{li2023clip, he2021transreid} incorporate camera information during training through specialized modules called \textit{Side Information Embeddings} (SIE), which combine camera ID data with Fourier embeddings to enhance overall accuracy. Yet, this limits the use of these methods in out-of-domain settings because the model weights are designed to accommodate only a specific number of camera IDs, as observed in their source domain. The work in \cite{song2025exploring} specifically tackles the issue of camera bias by proposing a post-hoc, black-box methodology for feature normalization based on camera IDs. This approach assumes that camera-bias factors can be estimated by calculating the mean and standard deviation of the target camera distribution, and that normalizing with these statistics can eliminate bias without requiring any adaptation.

\mypar{Test-Time Adaptation.~} TTA \cite{boudiaf2022parameter, iwasawa2021test, karani2021test, wang2020tent, yazdanpanah2022visual, ma2022test, you2021test, he2020self, yang2022dltta} focuses on adapting the source model to the target domain solely during inference. This presents two main advantages: (\emph{i}) it avoids iterative training, making it computationally efficient and easily deployable in online scenarios, and (\emph{ii}) it does not require target training data, which can enhance generalization to a wide range of target domains. TTA in person ReID, however, is  
underexplored compared to classification tasks. The method in \cite{adachi2024test} develops a batch-level entropy minimization strategy, building on \cite{wang2020tent}, to tackle domain shifts caused by image corruption. Anchor-ReID \cite{almansoori2023anchor} employs a Maximum Mean Discrepancy based loss between target samples and a memory bank of source data samples to address cross-dataset domain shift. These methods can be categorized as \textit{fully} TTA, as opposed to works such as \cite{han2022generalizable} where the source model architecture needs to be altered in a specific format, or 
\cite{zhou2021bayesian}, which need to modify the pre-training procedure. 

We observe that TTA methods designed for person ReID tasks are often adapted from classification techniques \cite{adachi2024test}, which tends to overlook important practical differences between these tasks. The SIE modules introduced by \cite{he2021transreid, li2023clip} operate solely on source data; moreover, models incorporating SIE modules cannot be applied to target data captured by different cameras, as their model weights are effectively “hard-coded” to the number of cameras in the source dataset. This limitation renders them impractical in our context. Lastly, while \cite{song2025exploring} proposes a straightforward approach to addressing camera bias, it relies on the oversimplified assumption that the bias parameters for each camera remain constant over time.

\section{Methodology}
\vspace{-8pt}

\mypar{Preliminary Notations:} Consider a model $f_\theta$ parameterized by a set of learnable weights $\theta$, which is pre-trained on a source dataset. We denote the target test dataset as $\DD = \{\DD_q, \DD_g\}$\footnote{For simplicity, we drop the use of standard subscripts $S$ and $T$ for source and target data/models, since any data mentioned in the following document is strictly drawn from the target dataset, unless specified otherwise}, with $\DD_q$ the query and $\DD_g$ the gallery. These are defined as $\DD_q = \{(\XX_i, c_i)\}_{i=1}^{N_q}$ and $\DD_g = \{(\XX_j, p_j, c_j)\}_{j=N_q +1}^{N_q+N_g}$ where $\XX \in \Real^{H \times W \times 3}$ is an RGB image of height $H$ and width $W$, $p$ is the specific person ID, and $c$ denotes the index of the camera that captured the image. $N_q$ and $N_g$ are the total size of the query and gallery datasets. Assuming there are a total of $n$ cameras, the set of camera IDs can be represented as $\CC = \{c_l\}_{l=1}^n$. Furthermore, to ensure consistency with the subsequent pseudocodes, we represent gallery indices as integers that follow those of the query data. In this way, the entire test dataset can be treated as a single, uniformly indexed set: the first $N_q$ indices correspond to the queries, and the next $N_g$ indices correspond to the gallery.

Having established these preliminaries, we denote as $\zz = f_\theta(\XX) \in \Real^d$ the image embedding that will be used for the matching task. Thus, we can readily use this model to encode all query and gallery images, storing their embeddings as two matrices $Q \in \Real^{N_q \times d}$ and $G \in \Real^{N_g \times d}$.

\subsection{Distance Based Adaptation Objective}
\begin{figure}[!thbp]
  \centering
  \begin{minipage}[]{0.48\textwidth}
    \centering
    \includegraphics[width=0.8\textwidth]{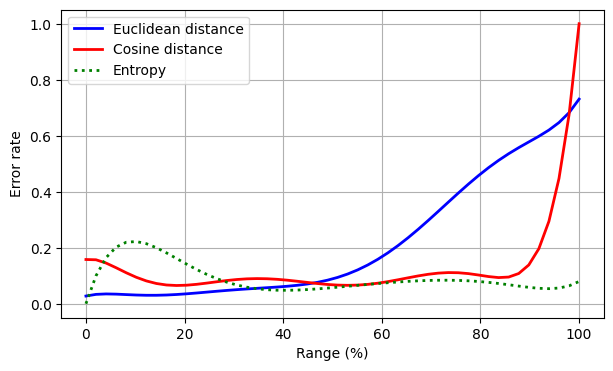} 
    \caption{Change in error rate vs nearest Euclidean and Cosine distance, and entropy, over a range of values. The error rate for Euclidean distance grows more uniformly over the range. In contrast, cosine distance grows abruptly for a high value, and with entropy the error rate behaves unpredictably}
    \label{fig:error_rate_curve}
  \end{minipage}
  \hfill
  \begin{minipage}[]{0.48\textwidth}
    \centering
    \centering
    \includegraphics[width=0.8\textwidth]{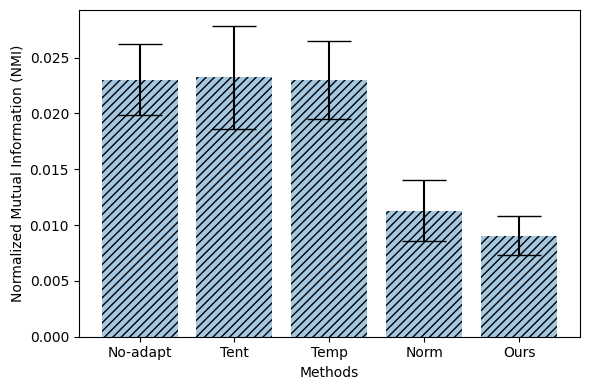} 
    \caption{Normalized Mutual Information (NMI) between feature clusters and Camera IDs. NMI serves as a measure of camera bias and our method demonstrates the lowest scores compared to existing methodologies.}
    \label{fig:nmi_plot}
    \vspace{0.6cm}
  \end{minipage}
  \vspace{-8pt}
\end{figure}

Typically, adaptation methods employ prediction entropy minimization based approaches \cite{wang2020tent, adachi2024test, hakim2024clipartt} or pseudo-label entropy minimization approaches \cite{goyal2022test, chen2022contrastive, wang2022towards}. \autoref{fig:error_rate_curve}, however, reveals the implicit flaw in an entropy minimization based approach. This curve was obtained by testing a CLIP-ReID \cite{li2023clip} model, trained on 5 cameras in the MSMT17 dataset and tested on an unseen camera. We observe that the error rate of a source pre-trained person ReID model \cite{li2023clip} behaves quite unpredictably and it can often increase when entropy is arbitrarily minimized. In case of cosine distance, the error rates steeply increase at a higher magnitude, but for lower values, the trend is more unpredictable. Therefore, among the common choices for distance measures, we find that error rate has the most stable progression with respect to changes in Euclidean distance. This behavior can be explained by the fact that a higher distance between query and gallery features implies a greater uncertainty that stray gallery points in between may lead to erroneous predictions.

The distance based score can therefore be computed as follows: $\Delta = \exp{(-\delta(\hat{\zz}, \tilde{G})/\tau)}$, where $\tau$ is a temperature parameter and $\delta(\hat{\zz}, \tilde{G})$ denotes the Euclidean distance matrix between a scaled-shifted feature batch $\hat\zz$ and the normalized gallery features $\tilde{G}$. We must note that the temperature scores in this case operates on unbounded logits, and therefore, a higher value of temperature is required (around 100-500) to prevent the gradients from vanishing due to precision limits. A simple Euclidean distance based objective, however, may result in the optimization collapsing on a trivial solution. To prevent this, a \textit{softmax} form of the distance, $H(\Delta) = -\log(\frac{\Delta}{\sum_{i=1}^{N_g} \Delta(:,i)})$ is used. Since we are only interested in the top-k closest points, we design a mask $M_{top\text{-}k}$ as follows:
\begin{equation}
 M_{top\text{-}k}(i,j) = 
 \begin{cases} 
 1, & \text{if $\Delta_{soft}(i,j)$ is in the top-k least values within $H(\Delta)(i,...)$} \\
 0, & \text{otherwise}
\end{cases}
\end{equation}

We therefore note that all three tensors $\Delta$, $\Delta_{soft}$ and $M_{top\text{-}k}$ have the same dimensionality: $B \times N_g$. With this, we may finally write the loss function as the mean of the sum of masked distances:
\begin{equation}
 \mathcal{L} = \frac{1}{B} \sum_{i=1}^{B} \sum_{j=1}^{N_g} H(\Delta)(i,j) \cdot M_{top-k}(i,j)
\end{equation} \label{eq:loss}

\subsection{Camera Bias and Optimization Parameters}
We start by examining the unique problem of camera bias, considering its nature and potential causes.  In the context of person ReID, camera bias refers to the model’s output features being skewed by background elements—often due to the presence of highly discriminative entities. This bias can arise because most ReID pipelines are trained using cross-entropy loss, which implicitly encourages the model to rely on any features that aid classification, including those unrelated to the target identity.

The true, \emph{unbiased} representation of the $i$-th image in the dataset can be denoted as $\zz_i^*$. We model of the camera-induced bias using a combination of scale and shift quantities as follows:
\begin{equation} \label{eq:bias_def}
 \zz_i = \alpha_{c_i}\zz_i^* + \beta_{c_i}
\end{equation}
where $c_i$ is the camera ID for the $i^{th}$ image. Based on this model, if the scale and shift parameters are found, then the unbiased representation can be recovered as
\begin{equation} \label{eq:bias_def}
 \zz_i^* = \frac{\zz_i - \beta_{c_i}}{\alpha_{c_i}}
\end{equation}

In our method, we define external learnable parameters $\Sigma(c_i)$ and $M(c_i)$, that use the camera ID $c_i$ as a key and store a vector of same dimensionality as $\zz_i$. Specifically, we use two distinct sets of parameters, each for performing scale and shift operations on the query and gallery features, separately. Thus, we have 4 groups of external parameters, $M_q$ and $\Sigma_q$ for the query and $M_g$ and $\Sigma_g$. We initialize the two sets of parameters: $M_q$ and $M_g$, and $\Sigma_q$ and $\Sigma_g$ with identical values, despite designating them as separate parameters. This is because we only make the set of parameters intended for the query scaling and shifting learnable i.e. $M_g$ and $\Sigma_g$ are frozen after initialization, since such an arrangement prevents the adaptation collapsing on a trivial solution. In addition, following \cite{wang2020tent}, the batchnorm parameters of the source model $f_{theta}$ are adapted at the test time.

The external parameters modify the output feature $\zz_i$ as follows:
\begin{equation} \label{eq:scale_shift}
    \hat{\zz}_i = \frac{\zz_i - M_q(c_i)}{\Sigma_q(c_i)}
\end{equation}
Eqs. (\ref{eq:bias_def}) and (\ref{eq:scale_shift}) show that as $M_q(c_i) \rightarrow \beta_{c_i}$ and $\Sigma_q(c_i) \rightarrow \alpha_{c_i}$, $\hat{\zz}_i$ tends towards the true representation $\zz^*_i$. Our primary goal towards the TTA therefore, is to estimate the optimal parameters such that a biased feature representation can be unbiased using post-hoc scale and shift operations. Further details and empirical evidence concerning the effects of camera bias has been provided in the Supplementary Material.

\begin{figure*}[!t]
 \centering
 \includegraphics[width=0.8\linewidth]{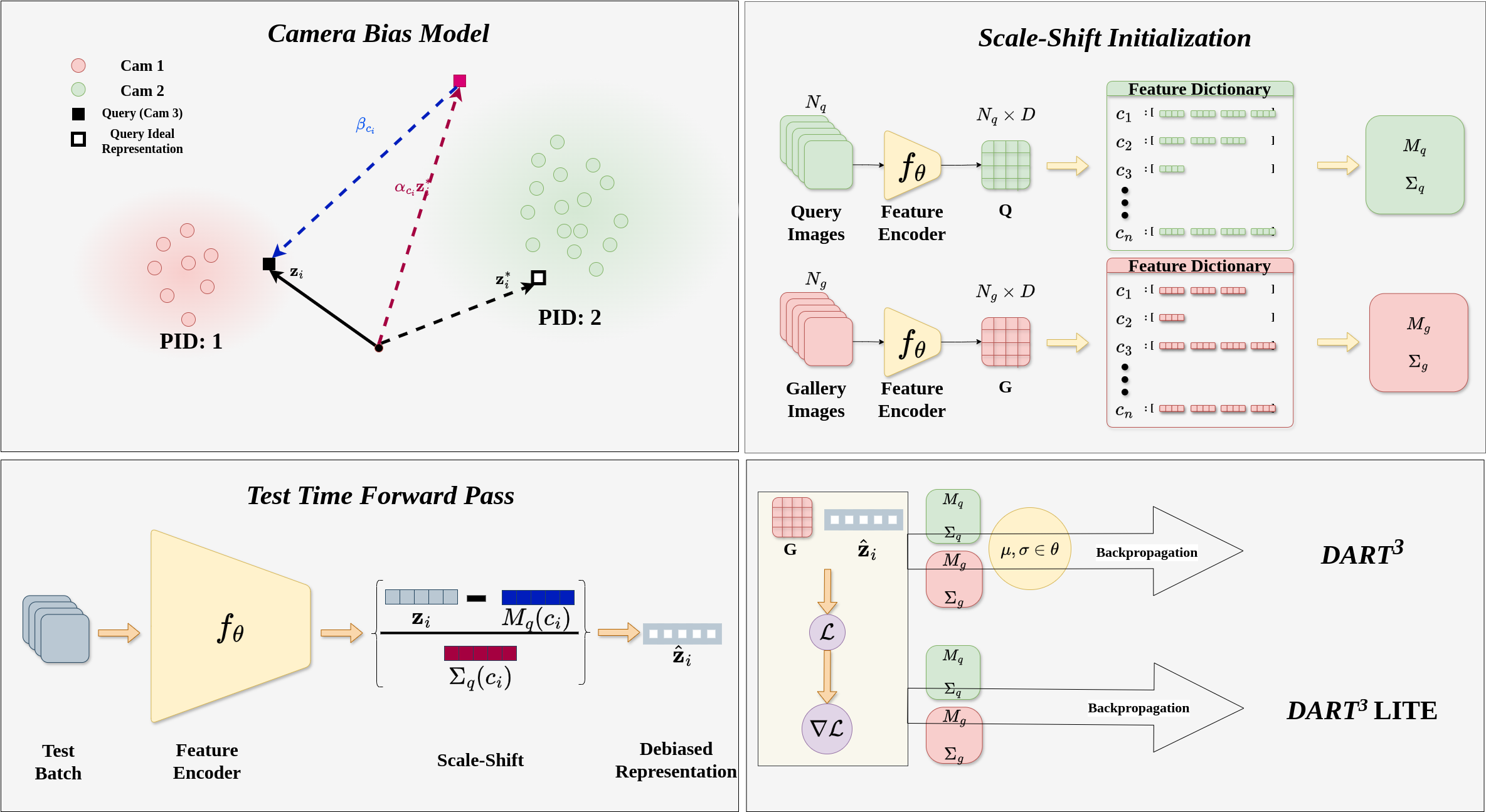}
 \caption{We present \method, a test time adaptation pipeline designed to mitigate camera bias in ReID models when exposed to unseen camera domains. We hypothesize that a true unbiased representation $\zz^*_i$ exists and can be estimated by scale and shift parameters. For the adaptation, these parameters can be initialized as the mean and standard deviation of features accumulated for a specific camera. Finally we show two variants of our method, \method and \method LITE, based on whether we treat the source model as a black-box entity.}
 \label{fig:method}
 \vspace{-8pt}
\end{figure*}

\subsection{Distance-Aware Retrieval Tuning at Test Time (\method)}

The proposed \method comprises of two key stages: (1) initialization, and (2) iteration over test batches. A complete summary of the methodology is provided in the Supplementary Material.

\mypar{Initialization: } Following \cite{song2025exploring} we explore camera based feature normalization for initializing the external scale and shift parameters. \autoref{fig:nmi_plot} shows that a camera normalization using a global camera mean and standard deviation is effective in removal of some of the camera bias. This plot was obtained by comparing the features from unseen camera samples from the MSMT17 dataset, for a CLIP-ReID model \cite{li2023clip} trained on only 5 cameras from the dataset. However, it is impractical to assume that a pre-computed global normalization statistic sufficiently models the bias, due to noise in real world data, and it is better to adapt these values to a better estimate for each batch of data. We therefore, initialize the external scale-shift parameters $M$ and $\Sigma$ by computing the normalization statistics for features belonging to each camera. Let us consider a specific camera $c \in C$, among the total set of cameras available in the test set, i.e. $C$. Thus, for all the query image encodings $\mathbf{z_i}$, for which $c_i = c$, we may find the mean and standard deviations of encodings as the vectors: $\mathbf{\mu}_{q,c} = \frac{1}{|D_q|_{c}} \sum_{i,c_i=c}^{N_q} \zz_i$ and $ \mathbf{\sigma}_{q,c} = \sqrt{\frac{1}{|D_q|_{c}} \sum_{i,c_i=c}^{N_q} (\mathbf{z_i} - \mathbf{\mu}_{q,c}) \odot (\mathbf{z_i} - \mathbf{\mu}_{q,c}) }$. Here, we use the notation $|D_q|_{c}$ to represent the number of query images from the camera $c$. The normalized feature can thus be represented as\footnote{The tilde mark ('$\text{  }\tilde{}\text{  }$') will be used exclusively to denote normalized quantities.}:
\begin{equation} \label{eq:norm}
 \tilde{\zz}_i = \frac{\zz_i - \mu_{q, c_i}}{\sigma_{q,c_i}}
\end{equation}

The mean and standard deviation can be computed for the gallery dataset as well, in a similar fashion.  The mean and variances for each camera $c \in C$ can thus be used to initialize the 4 external parameters as: (1) $M_q = \{\mu_{q,c}\text{ }|\text{ }c\in C\}$, (2) $M_g = \{\mu_{g,c}\text{ }|\text{ }c\in C\}$, (3) $\Sigma_q = \{\sigma_{q,c}\text{ }|\text{ }c\in C\}$ and (4) $\Sigma_g = \{\sigma_{g,c}\text{ }|\text{ }c\in C\}$. A detailed explanation why the normalized forms of the biased representation $\tilde{\zz}_i$ and the normalized form of the true representation $\tilde{\zz}_i^*$ are mathematically identical has been provided in the Supplementary Material.  

If the entire query dataset $\DD_q$ is available, we may encode the entire query as a preliminary step as well to compute the camera statistics for initializing $M_q$ and $\Sigma_q$. This step is useful where the query batch contains samples only from an unseen camera.  In cases where the domain shift involves a change in the location and all the cameras in the test dataset, we know that the query and gallery contains images from the same camera and we may assume that the statistics $M_q\approx M_g$ and $\Sigma_q \approx \Sigma_g$. \autoref{fig:method} summarizes the normalization step and the two possible configurations of our method, depending on whether the model batchnorm parameters are being trained. The computation of the camera statistics can be performed following algorithm, provided in the Supplementary Material.

\mypar{Iteration:} For each batch of test data, we compute the loss defined in \autoref{eq:loss}. Non-episodic mode of adaptation is chosen as the default setting for \method -- the initialized parameters are not reset after processing a batch. Initialized parameters are updated by the gradient of our loss $\nabla \mathcal{L}$.

\mypar{\method LITE:} It is possible to treat the source model to be a black-box completely in our method, by adapting only the external scale-shift parameters. Since the external parameters are only a set of vectors, initialized as the mean and standard deviation for a set of features, the number of additional learnable parameters only depends on the number of cameras in the dataset and in a real application it only ranges to about 30k to 50k parameters to adapt. This is a substantially smaller number of parameters to adapt, as compared to the total number of batchnorm parameters in the model, which makes this variant of our method "\textit{LITE}". The LITE mode offers comparable performance at a substantially low computational cost.

\vspace{-8pt}
\section{Results and Discussion}
\vspace{-8pt}
\subsection{Experimental Settings}

\mypar{Implementation Details:}
The default mode of operation of our proposed method is non-episodic, and all scores reported must be assumed to be for the non-episodic mode unless stated otherwise. We used a temperature value $\tau=100$ for the CLIP-ReID \cite{li2023clip} baseline and $\tau=200$ for the TransReID baseline \cite{he2021transreid}, as lower values lead to gradients diminishing beyond the precision limits. Only one optimization step per batch was found to be sufficient and the learning rate is set to 1e-4. The k-value for the optimization is chosen to be 3.


\mypar{Settings for New Camera in an Existing Network:} A practical use case of this method is where surveillance systems incorporate new camera installations in their systems. This implies that the new camera is essentially installed on the same network of cameras at a location (such as a park, campus, market, etc), but the precise location of the camera within the region under surveillance is different. We simulate this scenario by training ReID backbones on a subset of cameras from existing public datasets, and testing on images from camera installations unseen to the pretrained model. Three datasets: MSMT17 \cite{wei2018person} which is arguably the largest and the most challenging dataset, Duke-MTMC \cite{ristani2016performance} and Market-1501 \cite{zheng2015scalable}, have been used. Specifically, we use 4 distinct settings where the source model is trained on: (1) 5 cameras from MSMT17 and tested on the remaining 10, (2) 10 cameras from MSMT17 and tested on the remaining 5, (3) 5 cameras from Duke-MTMC and tested on the remaining 3 and finally (4) 5 cameras from Market-1501 and tested on the remaining camera. The scores reported are the weighted means for each remaining unseen cameras in the test dataset.

\mypar{Settings for a New Camera Network:} We also consider a situation where a source pretrained ReID backbone is used at a completely new location, evaluated on a new network of cameras. The evaluation strategy for this case is exactly identical to that of cross-dataset domain-shift experiments. As before, we choose MSMT17 \cite{wei2018person}, Duke-MTMC \cite{ristani2016performance} and Market-1501 \cite{zheng2015scalable} datasets for evaluation. The experiment protocol dictates that the source model pre-trained on one of these datasets, is directly tested on the other two datasets.



\subsection{Performance on New Camera in an Existing Network:}

We observe from \autoref{tab:coloc} that using TENT \cite{wang2020tent} almost always results in either the same or a slightly deteriorated performance as compared to the source model. This can be explained by drawing insights from \autoref{fig:error_rate_curve}, where we can see that error rates increase unpredictably when entropy in minimized. TEMP \cite{adachi2024test} improves the formulation of the entropy scores by employing a cosine similarity between the query and gallery features, but even this is only enough to improve the mean Average Precision scores by 0.5\%. Camera Normalization at the test time \cite{song2025exploring}, has the best scores among existing methods to tackle the problem of camera bias. Finally, we note that \method significantly improves the performance over existing methods. We present scores for two different versions of our method, (1) the standard version and (2) the lightweight "LITE" version..

\begin{table}[!t]
\centering
\caption{Performance of \method against SOTA methods to address camera bias at the test time for Person ReID. For each dataset, the stated number of cameras are used for training the backbones \cite{li2023clip, he2021transreid}, and images from the rest of the cameras are used to evaluate the methodologies.} \label{tab:coloc}
\resizebox{0.8\columnwidth}{!}{%
\begin{tabular}{c|c|cc|cc|cc|cc}
\toprule
\multirow{2}{*}{\textbf{Backbone}} & \multirow{2}{*}{\textbf{Method}} & \multicolumn{2}{c|}{\textbf{MSMT17 (5 cam)}} & \multicolumn{2}{c|}{\textbf{MSMT17 (10 cam)}} & \multicolumn{2}{c|}{\textbf{Duke-MTMC (5 cam)}} & \multicolumn{2}{c}{\textbf{Market-1501 (5 cam)}} \\
 & & \textit{mAP} & \textit{R1} & \textit{mAP} & \textit{R1} & \textit{mAP} & \textit{R1} & \textit{mAP} & \textit{R1} \\ \midrule
\multirow{7}{*}{TransReID \cite{he2021transreid}} & Upper Bound & 48.6 & 72.1 & 59.4 & 80.3 & 79.0 & 87.9 & 88.9 & 95.2 \\
\cmidrule{2-10}
& No-adapt & 34.4 & 41.3 & 46.7 & 62.7 & 66.7 & 77.8 & 85.2 & 93.1 \\
& TENT \cite{wang2020tent} & 32.8 & 40.0 & 44.6 & 60.6 & 65.6 & 77.1 & 84.5 & 92.6 \\
 & TEMP \cite{adachi2024test} & 33.8 & 41.1 & 46.3 & 61.9 & 65.8 & 76.5 & 84.4 & 92.7 \\
 & Norm \cite{song2025exploring} & 35.0 & 42.3 & 47.0 & 63.3 & 67.8 & 78.5 & 86.6 & 93.8 \\
 \rowcolor{gray!15}\cellcolor{white} & \method & 35.5 & 43.8 & 48.1 & 65.0 & 69.0 & 79.6 & 85.8 & 93.6 \\
 \rowcolor{gray!15}\cellcolor{white} & \method LITE & \textbf{36.2} & \textbf{44.4} & \textbf{49.8} & \textbf{67.1} & \textbf{70.1} & \textbf{80.8} & \textbf{86.8} & \textbf{93.7} \\ \midrule
\multirow{7}{*}{CLIP-ReID\cite{li2023clip}} & Upper Bound & 53.0 & 65.4 & 65.5 & 84.2 & 80.1 & 89.1 & 89.6 & 95.5 \\
\cmidrule{2-10}
& No-adapt & 42.1 & 50.0 & 54.0 & 69.2 & 70.6 & 82.0 & 85.7 & 94.0 \\
& TENT \cite{wang2020tent} & 41.8 & 49.8 & 53.1 & 68.3 & 70.7 & 82.0 & 85.7 & 93.9 \\
 & TEMP \cite{adachi2024test} & 42.4 & 50.3 & 54.7 & 70.0 & 70.8 & 82.0 & 85.9 & 94.5 \\
 & Norm \cite{song2025exploring} & 43.2 & 51.9 & 53.3 & 69.0 & 71.9 & 83.2 & 86.6 & 94.5 \\
 \rowcolor{gray!15}\cellcolor{white} & \method & \textbf{46.0} & \textbf{55.3} & 56.2 & 72.9 & 73.8 & 84.3 & 86.7 & 94.6 \\
 \rowcolor{gray!15}\cellcolor{white} & \method LITE & 45.9 & \textbf{55.3} & \textbf{57.2} & \textbf{74.0} & \textbf{73.9} & \textbf{84.5} & \textbf{86.9} & \textbf{94.8} \\ 
 \midrule
\end{tabular}}
\end{table}

\subsection{Performance on a New Camera Network:}

\autoref{tab:cross_dataset} shows that for TransReID \cite{he2021transreid}, the lighter version of our method functions unanimously better than the standard batch-norm adaptive version. This is likely due to the fact that a larger number of trainable parameters are more likely to experience overfit during the optimization, as compared to the LITE mode parameters which are only about ~10k to ~50k in number, depending on the dimensionality of the output of the model and the number of cameras in the target dataset. For CLIP-ReID \cite{li2023clip} however, this is not the case, although the LITE mode performing nearly identically as that of the standard mode. This may be attributed to the pretraining alignment of the CLIP model \cite{radford2021learning}, which implies that in general, output features are quite close to each other, spatially.

\begin{table}[!b]
\centering
\caption{Performance of \method against SOTA methods to address camera bias at a different location. The upper bounds essentially indicate supervised training on the specified datasets. We compare two standard backbones \cite{li2023clip, he2021transreid}, adapted using existing methodologies.} \label{tab:cross_dataset}
\resizebox{0.8\columnwidth}{!}{%
\begin{tabular}{c|l|cccc|cccc|cccc}
\toprule
\multirow[b]{3}{*}{\textbf{Backbone}} & \multirow[b]{3}{*}{\textbf{Method}} & \multicolumn{4}{c|}{\textbf{Source: MSMT17}} & \multicolumn{4}{c|}{\textbf{Source: Duke-MTMC}} & \multicolumn{4}{c}{\textbf{Souce: Market-1501}} \\ \cmidrule(l{2pt}r{2pt}){3-6}\cmidrule(l{2pt}r{2pt}){7-10} \cmidrule(l{2pt}r{2pt}){11-14}
 & & \multicolumn{2}{c}{\textbf{Duke-MTMC}} & \multicolumn{2}{c|}{\textbf{Market-1501}} & \multicolumn{2}{c}{\textbf{MSMT17}} & \multicolumn{2}{c|}{\textbf{Market-1501}} & \multicolumn{2}{c}{\textbf{MSMT17}} & \multicolumn{2}{c}{\textbf{Duke-MTMC}} \\ \cmidrule(l{2pt}r{2pt}){3-6}\cmidrule(l{2pt}r{2pt}){7-10} \cmidrule(l{2pt}r{2pt}){11-14}
 & & \textit{mAP} & \textit{R1} & \textit{mAP} & \textit{R1} & \textit{mAP} & \textit{R1} & \textit{mAP} & \textit{R1} & \textit{mAP} & \textit{R1} & \textit{mAP} & \textit{R1} \\ 
 \midrule
\multirow{7}{*}{TransReID \cite{he2021transreid}} & 
 Upper Bound & 80.5 & 89.7 & 87.0 & 94.6 & 55.9 & 77.2 & 87.0 & 94.6 & 55.9 & 77.2 & 80.5 & 89.7 \\
 \cmidrule{2-14}
& No-adapt & 54.3 & 70.8 & 42.9 & 68.2 & 15.2 & 37.5 & 38.2 & 65.2 & 15.6 & 37.0 & 47.2 & 64.9 \\
 & TENT & 52.3 & 69.3 & 40.2 & 65.4 & 10.7 & 26.8 & 36.9 & 62.8 & 10.5 & 25.4 & 45.2 & 63.4 \\
 & TEMP & 53.5 & 69.4 & 42.5 & 67.2 & 14.9 & 36.1 & 37.9 & 64.3 & 15.2 & 34.9 & 46.4 & 64.0 \\
 & Norm & 54.9 & 71.3 & 44.8 & 70.3 & 15.3 & 37.5 & 40.0 & 67.0 & 15.7 & 37.0 & 47.5 & 65.4 \\
 \rowcolor{gray!15}\cellcolor{white} & \method & 57.1 & 72.9 & 51.2 & 74.9 & 16.5 & 39.3 & 47.5 & \textbf{72.9} & \textbf{17.4} & 39.7 & 51.4 & 69.7 \\
 \rowcolor{gray!15}\cellcolor{white} & \method LITE & \textbf{57.8} & \textbf{73.9} & \textbf{51.6} & \textbf{75.8} & \textbf{16.8} & \textbf{40.4} & \textbf{47.6} & \textbf{72.9} & \textbf{17.4} & \textbf{39.8} & \textbf{52.6} & \textbf{71.3} \\
 \midrule
\multirow{7}{*}{CLIP-ReID \cite{li2023clip}} 
 & Upper Bound & 82.4 & 89.9 & 89.6 & 95.3 & 73.4 & 88.7 & 89.6 & 95.3 & 73.4 & 88.7 & 82.4 & 89.9 \\ 
\cmidrule{2-14}
& No-adapt & 57.0 & 73.8 & 50.3 & 75.6 & 21.1 & 47.4 & 43.3 & 70.8 & 23.2 & 49.4 & 50.1 & 68.8 \\
& TENT \cite{wang2020tent} & 56.8 & 73.8 & 49.7 & 74.8 & 17.9 & 41.2 & 42.9 & 69.8 & 20.4 & 45.2 & 50.0 & 68.6 \\
 & TEMP \cite{adachi2024test} & 57.1 & 73.8 & 50.4 & 75.4 & 21.4 & 48.0 & 43.5 & 70.8 & 23.5 & 50.0 & 50.2 & 68.9 \\
 & Norm \cite{song2025exploring} & 57.0 & 74.2 & 51.3 & 76.2 & 21.1 & 47.5 & 44.9 & 71.9 & 23.1 & 50.0 & 50.3 & 68.8 \\
 \rowcolor{gray!15}\cellcolor{white}&  \method & \textbf{59.5} & 76.4 & \textbf{56.7} & \textbf{81.9} & \textbf{23.7} & \textbf{51.2} & 50.3 & \textbf{76.5} & \textbf{25.7} & \textbf{53.5} & \textbf{54.0} & 72.2 \\
 \rowcolor{gray!15}\cellcolor{white} & \method LITE  & \textbf{59.5} & \textbf{76.6} & 56.1 & 79.7 & 23.1 & 50.5 & \textbf{50.9} & 76.2 & 25.1 & 52.7 & \textbf{54.0} & \textbf{72.3} \\ 
\bottomrule
\end{tabular}
}
\end{table}

We also document the performance of these methods in the source domain in \autoref{tab:self_dataset}. This experiment shows that camera bias is essentially similar in nature to an inductive bias, which means that under no domain shift, camera bias \textit{aids} the model performance, as opposed to diminishing it. We also observe that the results for our method is close to \cite{song2025exploring}, which validates our hypothesis that the $\alpha$ and $\beta$ biasing parameters are relatively constant for a non-complex (i.e. source-to-source) setting. The slight dip in performance in comparison to \cite{song2025exploring} might indicate that camera normalization gives a better estimate for the biasing parameters when there is no domain shift, as compared to our method, which likely enforces the errors instead of reducing them.

\begin{table*}[!t]
\centering
\begin{minipage}[t]{0.58\textwidth}
\scriptsize
\caption{Performance of \method against SOTA methods on source data using two standard backbones \cite{li2023clip, he2021transreid}.} \label{tab:self_dataset}
\vspace{0.3em}
\centering
\resizebox{\textwidth}{!}{%
\begin{tabular}{c|l|cc|cc|cc}
\toprule
\multirow[b]{2}{*}{\textbf{Backbone}} & \multirow[b]{2}{*}{\textbf{Method}} & \multicolumn{2}{c|}{\textbf{MSMT17}} & \multicolumn{2}{c|}{\textbf{Duke-MTMC}} & \multicolumn{2}{c}{\textbf{Market-1501}}\\
\cmidrule(l{2pt}r{2pt}){3-4} \cmidrule(l{2pt}r{2pt}){5-6} \cmidrule(l{2pt}r{2pt}){7-8}
 & & \textit{mAP} & \textit{R1} & \textit{mAP} & \textit{R1} & \textit{mAP} & \textit{R1} \\
\midrule
\multirow{6}{*}{TransReID \cite{he2021transreid}} 
& No-adapt & 66.4 & 84.7 & 80.5 & 89.7 & 87.0 & 94.6 \\
& TENT \cite{wang2020tent} & 64.7 & 83.1 & 80.0 & 89.2 & 86.2 & 93.9 \\
& TEMP \cite{adachi2024test} & 63.1 & 82.6 & 79.9 & 89.3 & 86.4 & 94.6 \\
& Norm \cite{song2025exploring} & 65.6 & 83.0 & 80.1 & 89.6 & 86.7 & 94.5 \\
\rowcolor{gray!15}\cellcolor{white} & \method & 61.9 & 77.8 & 78.7 & 88.9 & 85.7 & 93.7 \\
\rowcolor{gray!15}\cellcolor{white} & \method LITE & 65.4 & 77.0 & 79.4 & 89.3 & 86.9 & 94.6 \\
\midrule
\multirow{6}{*}{CLIP-ReID \cite{li2023clip}} 
& No-adapt & 73.4 & 88.7 & 82.4 & 89.9 & 89.6 & 95.3 \\
& TENT \cite{wang2020tent} & 69.3 & 85.4 & 82.4 & 90.1 & 89.4 & 95.1 \\
& TEMP \cite{adachi2024test} & 73.3 & 88.5 & 82.4 & 90.0 & 89.6 & 95.2 \\
& Norm \cite{song2025exploring} & 72.2 & 88.6 & 82.3 & 90.0 & 89.5 & 95.4 \\
\rowcolor{gray!15}\cellcolor{white} & \method & 72.3 & 88.5 & 81.2 & 89.9 & 89.5 & 95.1 \\
\rowcolor{gray!15}\cellcolor{white} & \method LITE & 72.1 & 88.6 & 81.2 & 90.0 & 89.6 & 95.4 \\
\bottomrule
\end{tabular}
}
\end{minipage}%
\hfill
\begin{minipage}[t]{0.38\textwidth}
\scriptsize
\caption{Ablation study: the effect of each design component in \method.}\label{tab:ablation}
\vspace{15pt}
\centering
\resizebox{\textwidth}{!}{%
\begin{tabular}{ccc|c|c}
\toprule
Source model & $M$, $\Sigma$ & $\mathcal{L}$ & mAP & R1 \\
\midrule
\checkmark &       &     &  42.1   &  50.0  \\
\checkmark & \checkmark &     &  43.2   &  51.9  \\
\checkmark &       & \checkmark &  41.7   &  50.0  \\
\checkmark & \checkmark & \checkmark &  45.9   &  55.3  \\
\bottomrule
\end{tabular}
}
\end{minipage}
\end{table*}


\begin{figure}[!htbp]
    \centering
    \begin{subfigure}[b]{0.22\textwidth}
        \includegraphics[width=\textwidth]{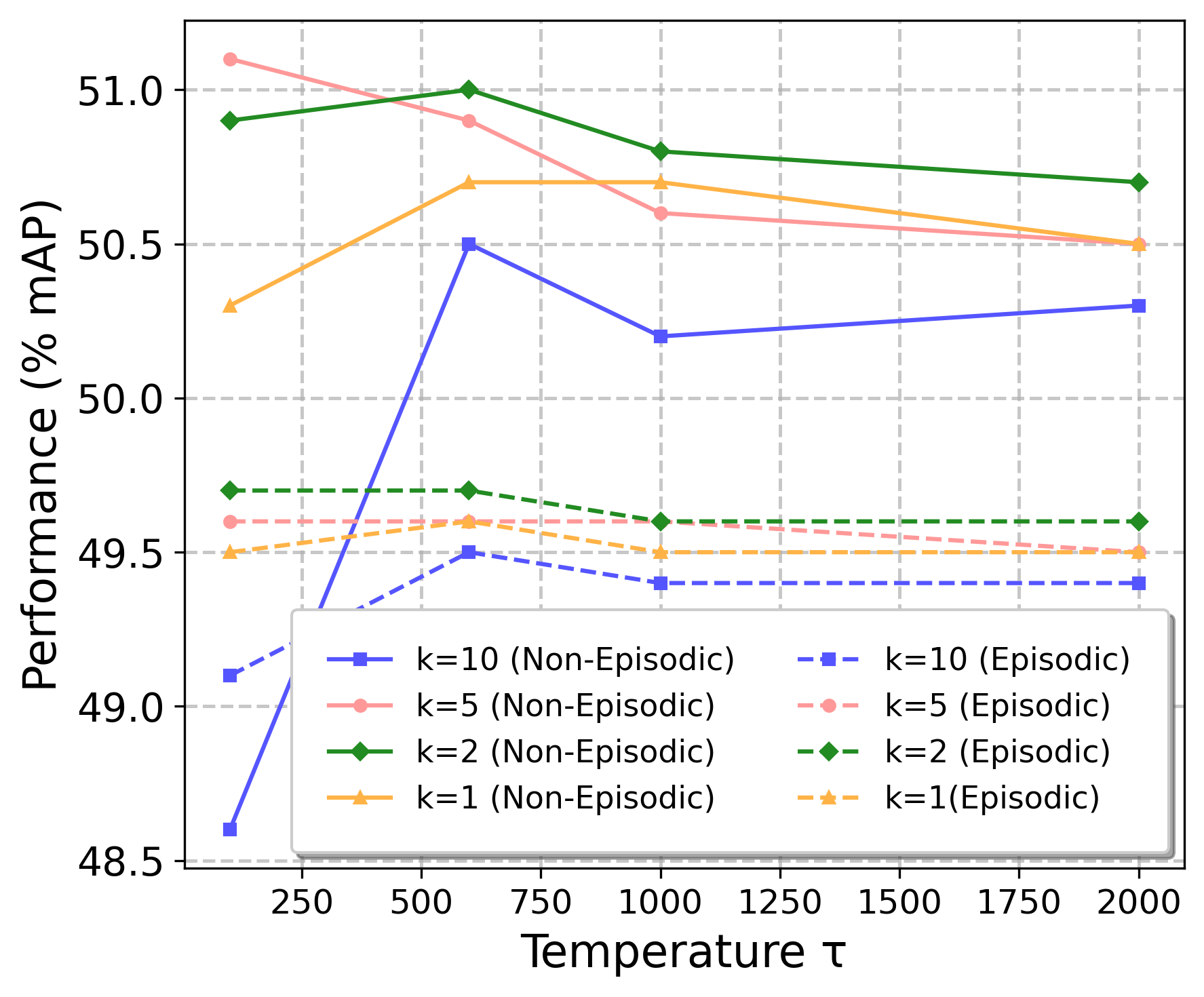}
        \caption{} \label{fig:abl_tau}
    \end{subfigure}
    \hfill
    \begin{subfigure}[b]{0.22\textwidth}
        \includegraphics[width=\textwidth]{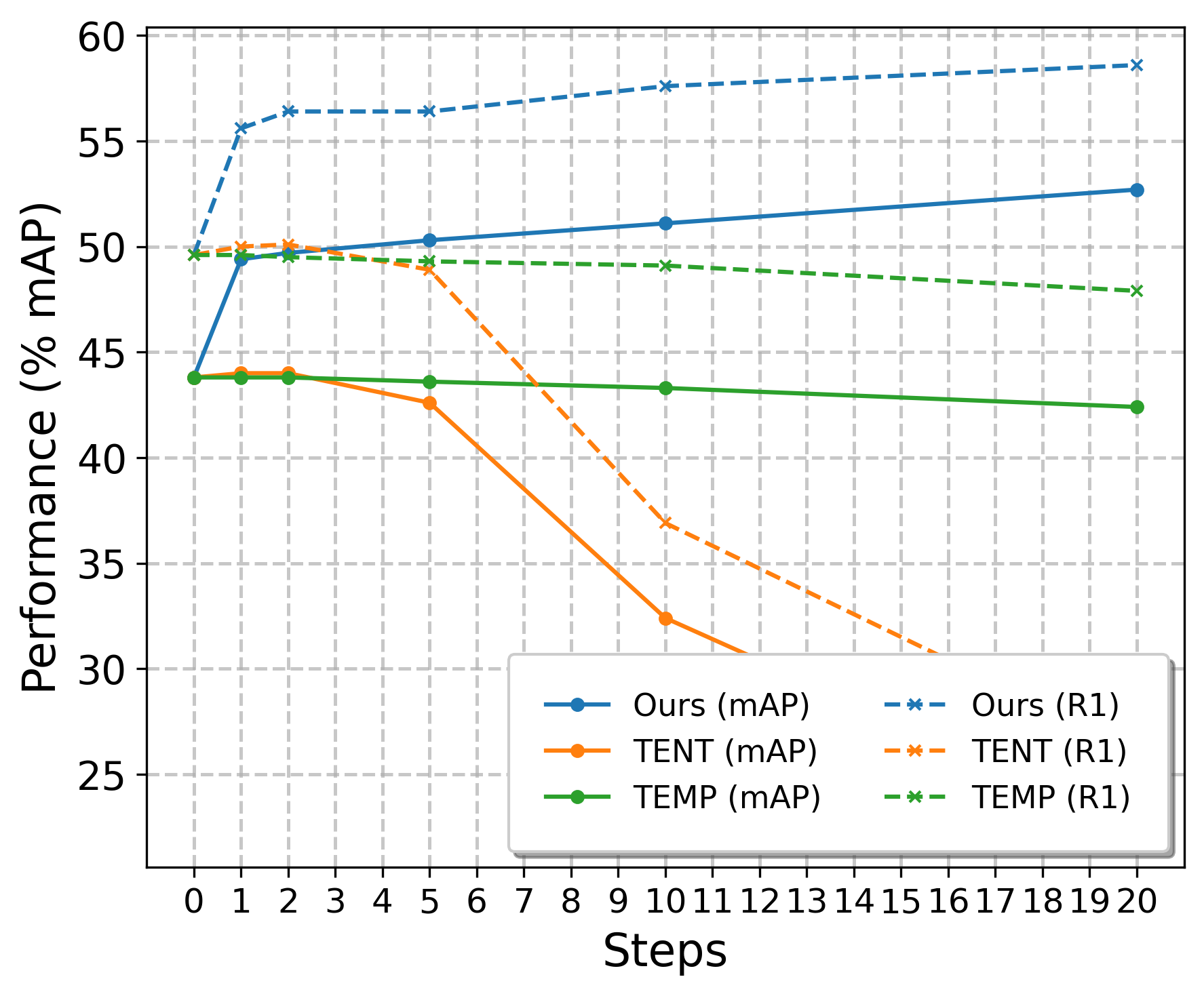}
        \caption{} \label{fig:abl_step_ne}
    \end{subfigure}
    \hfill
    \begin{subfigure}[b]{0.22\textwidth}
        \includegraphics[width=\textwidth]{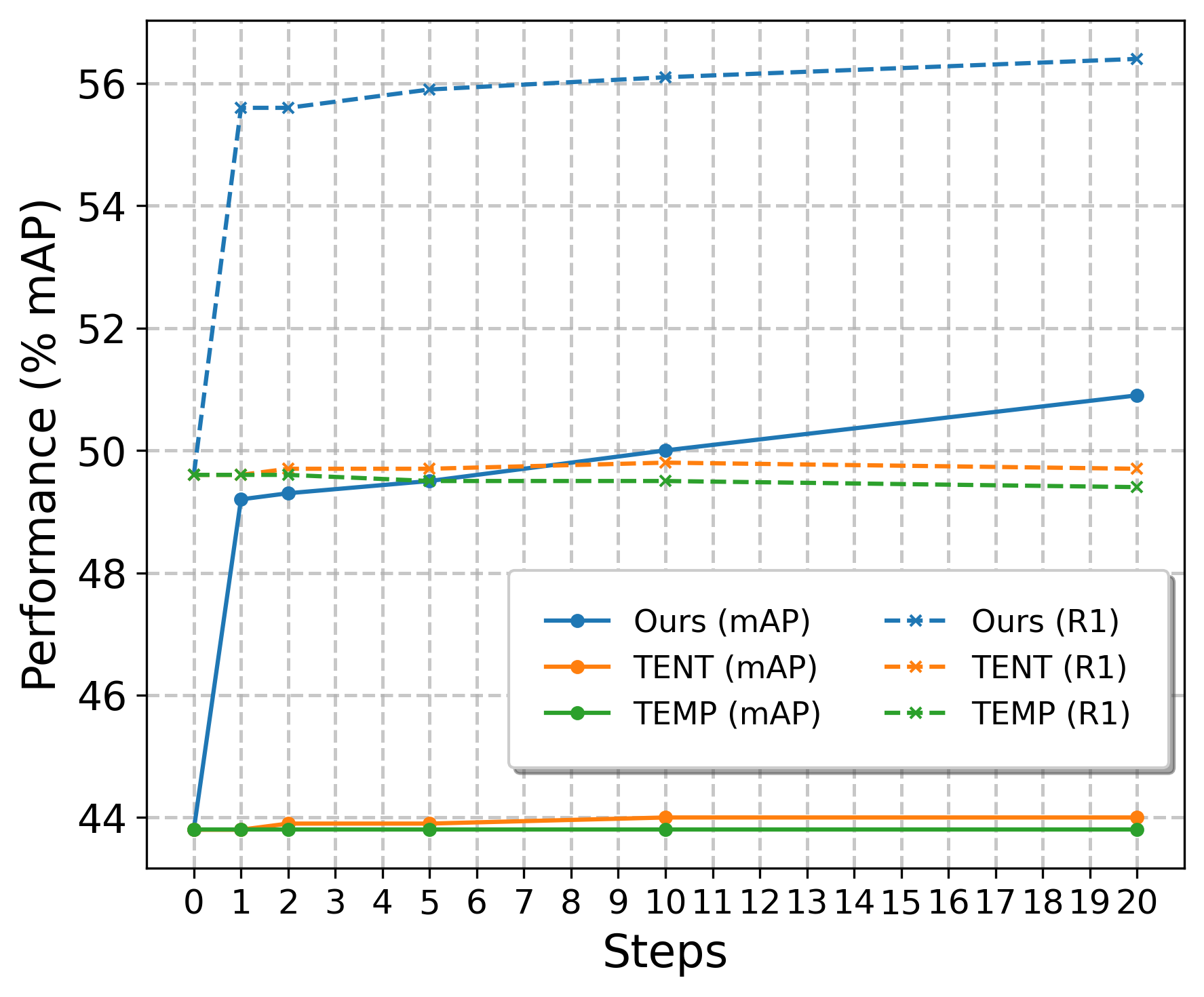}
        \caption{} \label{fig:abl_step_e}
    \end{subfigure}
    \hfill
    \begin{subfigure}[b]{0.3\textwidth}
        \includegraphics[width=\textwidth]{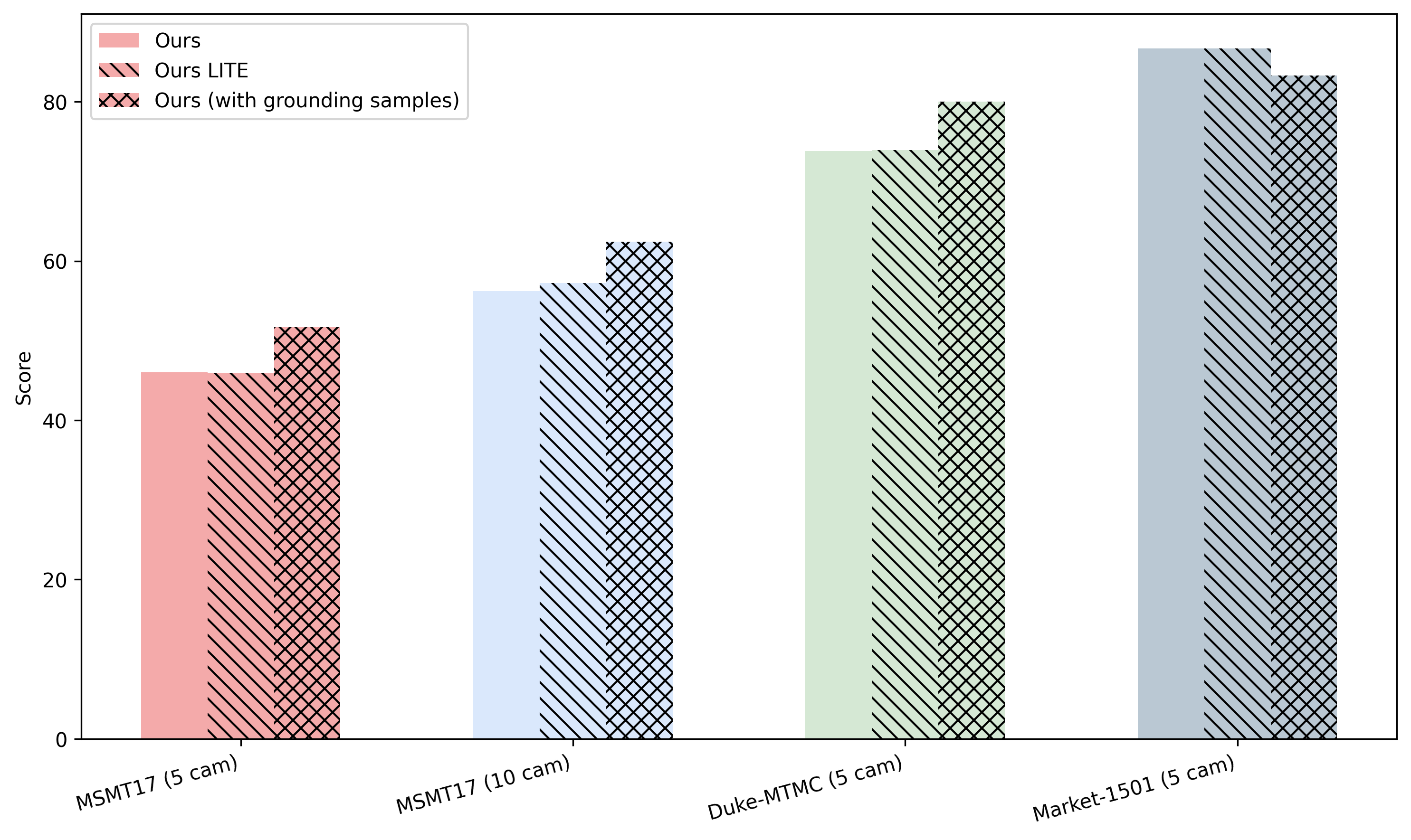}
        \caption{} \label{fig:wg_effect}
    \end{subfigure}
    \caption{Trends in performance (mAP) with respect to the value of $k$ and $\tau$ (a), and number of steps of optimization per batch for (b) non-episodic training and (c) episodic training. (d) Comparison of our method with grounding samples in a batch.}
\end{figure}


\subsection{Ablation Study}

In order to understand the contribution of each aspect of the methodology and the effect of varying each hyperparameter, we choose an unseen camera as the target data, from the MSMT17 \cite{wei2018person} dataset. We test out method, for a CLIP-ReID backbone, pretrained on 5 camera views in MSMT17 dataset.

\mypar{Component Study:} To compare the effect of the various design components of our algorithm, we perform an ablation study, as shown in \autoref{tab:ablation}. In theory, our method has only two components of note i.e. the external parameters $M$ and $\Sigma$ and the adaptation loss $\mathcal{L}$. The experiments are performed by using CLIP-ReID \cite{li2023clip} as the backbone, trained on 5 cameras and tested on the unseen cameras. Please note that the second row of the table is identical to Camera Normalization \cite{song2025exploring}, since we initialize the external parameters as the mean and standard deviation of the camera clusters. The ablation results validates our mathematical model for the bias, since the external parameters lead to the scale and shift correction of the output features, and without it, the loss function only enforces the errors that exist in the source model.

\mypar{Effect of $k$: } The selection of top-k Euclidean distances within a batch, is intended to soften the penalization of the loss over the batch of queries. For both episodic and non-episodic training, we find that a $k$ value between 1 to 5 performs the best, as can be seen in \autoref{fig:abl_tau}. However, choosing a larger $k$ value shows that the scores increase with an increasing temperature value. In general, it appears to be a good rule of thumb to choose a smaller $k$ value if a lower temperature is chosen, and vice-versa. We believe that this effect is seen largely due to lower temperature scores making the differences in distances larger, and as a result, it helps to have a smaller $k$ value, since it would prevent over-penalization.

\mypar{Effect of Softmax Temperature $\tau$:} Unlike classification confidence scores, the ReID model predicts spatial coordinates which are unbounded in nature. As a result, the temperature values used in our task is unusually large, in a range between 100-2000. For small temperature values, the resultant exponential value diminishes below the precision limit of the hardware used, thereby resulting in a collapse. This hyperparameter only seems to soften the effects of adaptation over multiple steps. From \autoref{fig:abl_tau}, we observe that a higher temperature score leads to a more stable trend in variations of performance over multiple steps and various values of $k$, but it leads to a lower peak, than a smaller value.

\mypar{Effect of Number of Steps: }
The number of steps is essentially the number of steps a batch of queries is optimized for. We find that the performance of our method consistently improves with the number of steps both in the episodic and non-episodic settings. \autoref{fig:abl_step_ne} and \autoref{fig:abl_step_e} demonstrates the two cases. We find that in comparison, entropy based methods TENT \cite{wang2020tent} and TEMP \cite{adachi2024test} drastically deteriorates over increasing number of steps in the non-episodic setting.

\mypar{Availability of Source Data at Test Time: Grounding Samples:} We find that if we break the homogeneity of the batches and add a small number of images from seen cameras i.e. source domain, we have a significant boost in the performance of the model. We call these additional samples "grounding samples" and these scores have also been reported for each method. While this method breaks the "source free" setup, it is not impractical to expect the existence and availability of source data in the scenario described, and the increase in performance justifies the departure from the fully source-free setup. The results have been depicted in the \autoref{fig:wg_effect}.


\mypar{Inference Time and the Effect of Batch Size}
\autoref{tab:inf_time} shows the effect of increasing batch size on the performance of the model. While methods like TENT \cite{wang2020tent} and TEMP \cite{adachi2024test} seems to require a larger batch for performing well, it also makes it a drawback, since larger batches can only be processed using more powerful computational devices. To this end, we observe that our method, delivers a stable performance across all batch sizes. While \method has a larger computational time, compared to the other method, we observe that \method LITE offers a significantly faster solution. This is due to the method requiring only one forward pass through the source model $f_\theta$, and the TTA essentially uses a batch of query features, instead of images. This approach, not only improves the computational time, but also makes the TTA algorithm both source-free and black-box in nature, thus enhancing its scope of applications. Our results are based on NVIDIA RTX A6000 GPU with approximately 50GB VRAM capacity.
\vspace{-10pt}

\begin{table}[!h]
\centering
\caption{Comparison of trends in performance as well as evaluation time taken per test set sample (in seconds). This experiment is performed using a source pretrained CLIP-ReID \cite{li2023clip}, trained on 5 cameras from MSMT17 dataset and tested on all test images from a single unseen camera.} \label{tab:inf_time}
\resizebox{0.9\columnwidth}{!}{%
\begin{tabular}{l|cc|cc|cc|cc|cc|cc}
\toprule
\multirow[b]{3}{*}{\textbf{Method}} & \multicolumn{12}{c}{\textbf{Batch Size}} \\ \cmidrule{2-13} 
 & \multicolumn{2}{c|}{\textit{1}} & \multicolumn{2}{c|}{\textit{2}} & \multicolumn{2}{c|}{\textit{8}} & \multicolumn{2}{c|}{\textit{16}} & \multicolumn{2}{c|}{\textit{32}} & \multicolumn{2}{c}{\textit{64}} \\ \cmidrule(l{2pt}r{2pt}){2-3}\cmidrule(l{2pt}r{2pt}){4-5}\cmidrule(l{2pt}r{2pt}){6-7}\cmidrule(l{2pt}r{2pt}){8-9}\cmidrule(l{2pt}r{2pt}){10-11}\cmidrule(l{2pt}r{2pt}){12-13}
 & \textbf{mAP} & \multicolumn{1}{c|}{\textbf{Time}} & \textbf{mAP} & \multicolumn{1}{c|}{\textbf{Time}} & \textbf{mAP} & \multicolumn{1}{c|}{\textbf{Time}} & \textbf{mAP} & \multicolumn{1}{c|}{\textbf{Time}} & \textbf{mAP} & \multicolumn{1}{c|}{\textbf{Time}} & \textbf{mAP} & \textbf{Time} \\ \midrule
No-adapt & 43.8 & 0.110 & 43.8 & 0.030 & 43.8 & 0.004 & 43.8 & 0.002 & 43.8 & 0.001 & 43.8 & 0.0003 \\
Norm \cite{song2025exploring} & 47.6 & 0.112 & 47.4 & 0.032 & 47.2 & 0.004 & 47.1 & 0.002 & 47.2 & 0.001 & 47.2 & 0.001 \\
\midrule
TENT \cite{wang2020tent} & - & - & 13.6 & 0.180 & 20.4 & 0.057 & 28.5 & 0.049 & 39.2 & 0.042 & 43.8 & 0.038 \\
TEMP \cite{adachi2024test} & - & - & 32.0 & 0.181 & 44.1 & 0.059 & 44.3 & 0.049 & 44.2 & 0.043 & 43.9 & 0.039 \\
\rowcolor{gray!15} \method & - & - & \textbf{49.1} & 0.408 & \textbf{49.6} & 0.116 & \textbf{49.3} & 0.077 & \textbf{49.0} & 0.058 & \textbf{48.5} & 0.047 \\
\rowcolor{gray!15} \method LITE & 47.6 & 0.136 & 47.6 & 0.092 & 47.7 & 0.047 & 47.6 & 0.046 & 47.6 & 0.045 & 47.6 & 0.043 \\ 
\bottomrule
\end{tabular}}
\end{table}

\vspace{-4pt}
\section{Limitations and Conclusion}
\vspace{-4pt}
We identify two main drawbacks of \method. First, the method has several tunable hyperparameters. In a real-world setting, despite our best efforts to quantify the effects of the hyperparameters and find optimal values, the availability of a validation set is necessary to determine the best hyper-parameters for a specific target dataset. Second, we find that \method is unable to improve the performance of the source model on the source distribution. Furthermore, despite improving the performance, our method does not inherently improve the properties of the baseline model in any way, and only uses a \textit{side-objective} to gain performance when exposed to out-of-domain distributions. Despite these limitations, our \method approach is tailored for person ReID under camera-induced domain shifts. Unlike conventional entropy-based objectives, our method leverages a distance-based loss that better aligns with the retrieval dynamics of ReID. By introducing external scale and shift parameters, we effectively mitigate camera bias without modifying the source model or requiring access to source data. Extensive experiments across multiple datasets and backbones indicates that our approach achieves consistent performance gains, including in black-box settings. This makes our method both practically viable and widely deployable in real-world surveillance systems. We aim to address the limitations of our approach in the future works.

\bibliographystyle{plain}
\bibliography{refs}

\newpage
\section*{Supplementary Material for: \\ \method: Leveraging Distance for Test Time Adaptation in Person Re-Identification}

\appendix

\section{Camera Bias Leading to Domain Shift Problems} \label{sec:app:bias}

ReID systems typically rely on a network of cameras to perform the identification task. A model trained to perform well on data derived from a network of cameras typically performs quite well in traditional ReID settings \cite{li2023clip}. However, the performance of these models tend to deteriorate as soon as they are introduced to video data from new cameras. This is typically due to identity features being biased against specific unseen cameras. \autoref{tab:app:exp2} shows this effect on two state-of-the-art models for traditional person ReID tasks, TransReID \cite{he2021transreid} and CLIP-ReID \cite{li2023clip}.

\begin{table}[!h]
\centering
\resizebox{0.7\columnwidth}{!}{%
\begin{tabular}{|c|c|c|cc|cc|}
\midrule
\multirow{2}{*}{\textbf{\#Images}} &
 \multirow{2}{*}{\textbf{\#Camera IDs}} &
 \multirow{2}{*}{\textbf{\#PIDs}} &
 \multicolumn{2}{c|}{\textbf{TransReID}} &
 \multicolumn{2}{c|}{\textbf{CLIP-ReID}} \\ \cmidrule{4-7} 
 & & & \textit{mAP} & \textit{R1} & \textit{mAP} & \textit{R1} \\ \midrule
32621 & 15 & 1041 & 65.0 & 84.0 & 73.2 & 88.8 \\
8245 & 5 & 998 & 34.4 & 41.3 & 42.1 & 50.0 \\
7937 & 15 & 934 & 48.6 & 72.1 & 53.0 & 65.4 \\ 
\bottomrule
\end{tabular}%
}
\vspace{3.0mm}
\caption{Performance of TransReID \cite{he2021transreid} and CLIP-ReID \cite{li2023clip} for  unseen camera domains within the MSMT17 dataset. This is done through reducing the training dataset to use data from only 5 camera IDs and 15 camera IDs with the same relative number of training images.}
\label{tab:app:exp2}
\end{table}

\begin{figure*}[!ht]
 \centering
 \includegraphics[width=\linewidth]{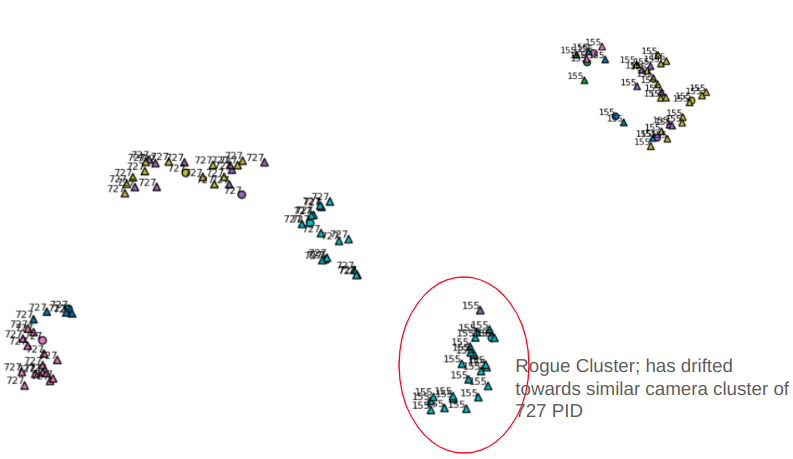}
 \caption{UMAP representation of two person ID clusters from Market-1501. Features are from CLIP-ReID \cite{li2023clip} pre-trained on MSMT17. The triangular data points represent gallery while circular ones represent queries. Colours represent camera ID capturing the image. Notice how one particular cluster from ID 155, drifts towards 727 based on camera ID. Since a simple spatial distance is used for evaluation and inference, this behavior can significantly affect model performance.}
 \label{fig:app:bias}
\end{figure*}

Notice how the performance drops drastically with roughly the same number of training images, for the second and third row where the models were trained on samples from only 5 and all 15 cameras respectively. This is a direct outcome of output representations being more strongly biased towards camera installations, instead of person identities. The effect can be seen visually in the following \autoref{fig:app:bias}. While the surrounding clusters are formed based on the person ID, as in the image, we observe that one particular cluster of data points has drifted close to another of differing person IDs, but having the same source camera.

The cause for camera bias is essentially the image background. This is because a ReID model does not have an intrinsic property to determine the installation of the source camera, unless the information is provided beforehand \cite{he2021transreid}, or was trained to match the image data captures with the source cameras. Therefore, while a ReID model prediction is ideally expected to be invariant to the backgrounds of the input images, state-of-the-art models cannot perfectly satisfy this criterion, when exposed to out-of-domain data -- in this case, data from a new camera. This phenomenon, as described in \autoref{fig:app:bias}, therefore, leads to two hypotheses about the nature of the bias: (1) numerically, the camera bias must be relatively constant, for images pertaining to a specific source camera, and (2) mathematically, camera bias has the nature of a vector shift or more generally scale and shift parameter. With respect to the output feature representations from a model, camera bias is therefore likely to be additive but we assume the most general case, where components of the bias are both additive and multiplicative, to develop a formalism.
Camera bias is assumed to be constant since this term is dependent only on the camera ID $c_i$, which is fixed for all image features corresponding to the specific camera ID $c_i$. Furthermore,  common ReID models typically operate with features in the Euclidean space \cite{li2023clip, he2021transreid}. This means that the features $\zz_i$ and $\zz_i$ exist in a $d$-dimensional Euclidean space, and the bias can essentially be modeled as scale and shift between these two points, thus assuming the $ax + b$ form.

\section{Methodology (Extended)}

\mypar{Preliminary Notations:} Consider a model $f_\theta$ parameterized by a set of learnable weights $\theta$, which is pre-trained on a source dataset. We denote the target test dataset as $\DD = \{\DD_q, \DD_g\}$\footnote{For simplicity, we drop the use of standard subscripts $S$ and $T$ for source and target data/models, since any data mentioned in the following document is strictly drawn from the target dataset, unless specified otherwise}, with $\DD_q$ the query and $\DD_g$ the gallery. These are defined as $\DD_q = \{(\XX_i, c_i)\}_{i=1}^{N_q}$ and $\DD_g = \{(\XX_j, p_j, c_j)\}_{j=N_q +1}^{N_q+N_g}$ where $\XX \in \Real^{H \times W \times 3}$ is an RGB image of height $H$ and width $W$, $p$ is the specific person ID, and $c$ denotes the index of the camera that captured the image. $N_q$ and $N_g$ are the total size of the query and gallery datasets. Assuming there are a total of $n$ cameras, the set of camera IDs can be represented as $\CC = \{c_l\}_{l=1}^n$. Furthermore, to ensure consistency with the subsequent pseudocodes, we represent gallery indices as integers that follow those of the query data. In this way, the entire test dataset can be treated as a single, uniformly indexed set: the first $N_q$ indices correspond to the queries, and the next $N_g$ indices correspond to the gallery.

Having established these preliminaries, we denote as $\zz = f_\theta(\XX) \in \Real^d$ the image embedding that will be used for the matching task. Thus, we can readily use this model to encode all query and gallery images, storing their embeddings as two matrices $Q \in \Real^{N_q \times d}$ and $G \in \Real^{N_g \times d}$.

\subsection{Test Time Adaptation: TENT \cite{wang2020tent} and TEMP \cite{adachi2024test}}
We dedicate this section to introducing TTA and two standard TTA techniques. Typically, TENT \cite{wang2020tent} based methodologies employ variations on entropy, as the adaptation objective at the test time. For a probability prediction $p(x)$ by a model, given an input image X, the entropy can be computed as:
\begin{equation}
 H(x) = - \sum_{i=1}^B p(\mathbf{X_i}) \log p(\mathbf{X_i})
\end{equation}

This equation describes the entropy for a batch of images with a size $B$. TENT \cite{wang2020tent} selects the scale and shift i.e. the BatchNorm parameters of the model and makes them trainable to minimize this loss. This leads to the benefit that the model does not lose its original optimization, and the model can easily be reverted back to its unadapted stage as needed. A similar methodology has been adopted recently by TEMP \cite{adachi2024test}, where the authors identify that the ReID models, unlike classification pipelines, do not provide representations that are bounded class probabilities. Instead, these features are mapped in terms of some spatial coordinates and predictions are made based on some spatial distance measures (e.g., Euclidean or Cosine distance). The authors argue that a form of entropy may be derived using the cosine distance between a batch of query features and all the gallery features, and computing the softmax on the top-k values of cosine distance. This is computed as:
\begin{equation}
 s_{ij} = \frac{\zz_i^q\zz_j^g}{||\zz_i^q||_2 ||\zz_j^g||_2}
\end{equation}
for query and gallery features $\zz_i^q$ and $\zz_j^g$, and thereafter, the entropy is given as:
\begin{equation}
 H_i = \sum_{j=1}^k \hat{p}_{ij} \log \hat{p}_{ij}
\end{equation}
where $\hat{p}_{ij} = \frac{\exp(s_{ia_{ij}})}{\sum_{j'=1}^k \exp(s_{ia_{ij'}})}$, $a_{ij} \in \{1,2,3, \dots n\}$. This entropy is the test time objective function to be minimized by backward propagation. The source model weights are left frozen while the scale-shift parameters i.e. batchnorm parameters are adapted using this method.

\subsection{Distance Based Adaptation Objective}

\begin{figure}[htbp]
  \centering
  \begin{minipage}[b]{0.48\textwidth}
    \centering
    \includegraphics[width=\textwidth]{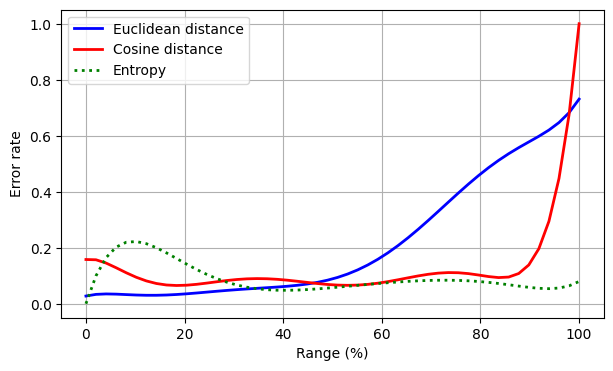} 
    \caption{Change in error rate vs nearest Euclidean and Cosine distance, and entropy, over a range of values. The error rate for Euclidean distance grows more uniformly over the range. In contrast, cosine distance grows abruptly for a high value, and with entropy the error rate behaves unpredictably}
    \label{supp:fig:error_rate_curve}
  \end{minipage}
  \hfill
  \begin{minipage}[b]{0.48\textwidth}
    \centering
    \begin{tabular}{lcc}
      \toprule
      \textbf{Distance} & \textbf{mAP} & \textbf{R1} \\
      \midrule
    Euclidean Distance  & 49.6 & 55.8 \\
    Cosine Distance  & 47.6 & 54.6 \\
      \bottomrule
    \end{tabular}
    \vspace{40pt}
    \captionof{table}{Comparison in performance when $\delta$ is chosen to be Euclidean and Cosine distances. We use Camera ID 0 samples on a CLIP-ReID model trained on 5 other cameras, as our unseen evaluation data. Our findings supports our choice of distance function.}\label{supp:tab:dists} 
    
  \end{minipage}
\end{figure}

Typically, adaptation methods employ prediction entropy minimization based approaches \cite{wang2020tent, adachi2024test, hakim2024clipartt} or pseudo-label entropy minimization approaches \cite{goyal2022test, chen2022contrastive, wang2022towards}. \autoref{supp:fig:error_rate_curve}, however, reveals the limitation in an entropy minimization based approach. This curve was obtained by testing a CLIP-ReID \cite{li2023clip} model, trained on 5 cameras in the MSMT17 dataset and tested on an unseen camera. We observe that the error rate of a source pre-trained person ReID model \cite{li2023clip} behaves quite unpredictably and it can often increase when entropy is arbitrarily minimized. In case of cosine distance, the error rates steeply increase at a higher magnitude, but for lower values, the trend is more unpredictable. Therefore, among the common choices for distance measures, we find that error rate has the most stable progression with respect to changes in Euclidean distance. This behavior can be explained by the fact that a higher distance between query and gallery features implies a greater uncertainty that stray gallery points in between may lead to erroneous predictions.

\paragraph{Proposition 1:} \emph{Let $x_q$ and $x_g$ be two feature vectors in a Euclidean space $\mathbb{R}^d$, representing a query and a gallery sample respectively, both belonging to the same class. Let $u: \mathbb{R}_{\geq 0} \rightarrow \mathbb{R}_{\geq 0}$ be a function that models the uncertainty in classification as a function of the Euclidean distance $d(x_q, x_g) = \|x_q - x_g\|_2$. Then, it is intuitive to assume that $u$ satisfies the following properties}:

\begin{enumerate}
    \item \textbf{Boundary condition:} $u(0) = 0$ \\
    \textit{Interpretation:} If the query and gallery embeddings are identical, the prediction is made with absolute certainty.
    
    \item \textbf{Monotonicity:} $u$ is monotonically increasing, i.e., if $d_1 < d_2$, then $u(d_1) < u(d_2)$. \\
    \textit{Interpretation:} As the distance between query and gallery increases, the uncertainty in prediction also increases.
\end{enumerate}

\paragraph{Corollary 1 (Certainty at zero distance).} If $x_q = x_g$, then $d(x_q, x_g) = 0$ and the uncertainty $u(0) = 0$.

\paragraph{Corollary 2 (Uncertainty grows with distance).} For any two distances $d_1, d_2 \in \mathbb{R}_{\geq 0}$ such that $d_1 < d_2$, we have $u(d_1) < u(d_2)$.

\paragraph{Discussion.} While $u$ is not derived from a probabilistic model and does not represent entropy in a formal sense, it serves as a conceptual proxy for uncertainty in tasks involving metric-based retrieval or classification. The assumption that uncertainty grows with increasing distance is consistent with the intuition that larger distances increase the likelihood of a closer but incorrect class neighbor, thereby degrading classification reliability.

Our distance based score is, therefore, computed as follows: $\Delta = \exp{(-\delta(\hat{\zz}, \tilde{G})/\tau)}$, where $\tau$ is a temperature parameter and $\delta(\hat{\zz}, \tilde{G})$ denotes the Euclidean distance matrix between a scaled-shifted feature batch $\hat\zz$ and the normalized gallery features $\tilde{G}$. While our choice of distance measure is supported by the empirical observation in \autoref{supp:fig:error_rate_curve}, we perform experimentation using both distances and our experiments reveal that Euclidean distance based loss leads to a better performance (Table~\ref{supp:tab:dists}). We must note that the temperature scores in this case operates on unbounded logits, and therefore, a higher value of temperature is required (around 100-500) to prevent the gradients from vanishing due to precision limits. A simple Euclidean distance based objective, however, may result in the optimization collapsing on a trivial solution. To prevent this, a \textit{softmax} form of the distance, $H(\Delta) = -\log(\frac{\Delta}{\sum_{i=1}^{N_g} \Delta(:,i)})$ is used. Since we are only interested in the top-k closest points, we design a mask $M_{top\text{-}k}$ as follows:
\begin{equation}
 M_{top\text{-}k}(i,j) = 
 \begin{cases} 
 1, & \text{if $\Delta_{soft}(i,j)$ is in the top-k least values within $H(\Delta)(i,...)$} \\
 0, & \text{otherwise}
\end{cases}
\end{equation}

We therefore note that all three tensors $\Delta$, $\Delta_{soft}$ and $M_{top\text{-}k}$ have the same dimensionality: $B \times N_g$. With this, we may finally write the loss function as the mean of the sum of masked distances:
\begin{equation}
 \mathcal{L} = \frac{1}{B} \sum_{i=1}^{B} \sum_{j=1}^{N_g} H(\Delta)(i,j) \cdot M_{top-k}(i,j)
\end{equation} \label{supp:eq:loss}

\subsection{Camera Bias and Optimization Parameters}
We start by examining the unique problem of camera bias, considering its nature and potential causes.  In the context of person ReID, camera bias refers to the model’s output features being skewed by background elements—often due to the presence of highly discriminative entities. This bias can arise because most ReID pipelines are trained using cross-entropy loss, which implicitly encourages the model to rely on any features that aid classification, including those unrelated to the target identity.

The true, \emph{unbiased} representation of the $i$-th image in the dataset can be denoted as $\zz_i^*$. We model the camera-induced bias using a combination of scale and shift quantities as follows:
\begin{equation}
 \zz_i = \alpha_{c_i}\zz_i^* + \beta_{c_i}
\end{equation}
where $c_i$ is the camera ID for the $i^{th}$ image. Based on this model, if the scale and shift parameters are found, then the unbiased representation can be recovered as
\begin{equation} \label{eq:bias_def}
 \zz_i^* = \frac{\zz_i - \beta_{c_i}}{\alpha_{c_i}}
\end{equation}

In our method, we define external learnable parameters $\Sigma(c_i)$ and $M(c_i)$, that use the camera ID $c_i$ as a key and store a vector of same dimensionality as $\zz_i$. Specifically, we use two distinct sets of parameters, each for performing scale and shift operations on the query and gallery features, separately. Thus, we have 4 groups of external parameters, $M_q$ and $\Sigma_q$ for the query and $M_g$ and $\Sigma_g$. We initialize the two sets of parameters: $M_q$ and $M_g$, and $\Sigma_q$ and $\Sigma_g$ with identical values, despite designating them as separate parameters. This is because we only make the set of parameters intended for the query scaling and shifting learnable i.e. $M_g$ and $\Sigma_g$ are frozen after initialization, since such an arrangement prevents the adaptation collapsing on a trivial solution. In addition, following \cite{wang2020tent}, the batchnorm parameters of the source model $f_{\theta}$ are adapted at the test time.

The external parameters modify the output feature $\zz_i$ as follows:
\begin{equation} \label{eq:scale_shift}
    \hat{\zz}_i = \frac{\zz_i - M_q(c_i)}{\Sigma_q(c_i)}
\end{equation}
Eqs. (\ref{eq:bias_def}) and (\ref{eq:scale_shift}) show that as $M_q(c_i) \rightarrow \beta_{c_i}$ and $\Sigma_q(c_i) \rightarrow \alpha_{c_i}$, $\hat{\zz}_i$ tends towards the true representation $\zz^*_i$. Our primary goal towards the TTA therefore, is to estimate the optimal parameters such that a biased feature representation can be unbiased using post-hoc scale and shift operations.

\subsection{Camera Based Feature Normalization as a Naive solution} \label{supp:sec:camnorm}
We want to show that a \textit{camera normalized} representation $\tilde{\zz}_i$ is essentially equivalent to the true representation $\zz_i^*$. Analyzing the definitions for mean and standard deviation, we find that:

\begin{equation} \label{supp:eq:trumean}
\begin{split}
 \mathbf{\mu}_{q,c} &= \frac{1}{|D_q|_{c}} \sum_{i,c_i=c}^{N_q} \zz_i \\
 &= \frac{1}{|D_q|_{c}} \sum_{i,c_i=c}^{N_q} (\alpha_{c_i}\zz_i^* + \beta_{c_i}) \\
 &= \beta_{c_i} + \frac{1}{|D_q|_{c}} \alpha_{c_i}\sum_{i,c_i=c}^{N_q} (\zz_i^*) \\
 &= \alpha_{c_i}\mathbf{\mu}_{q,c}^* + \beta_{c_i}
\end{split}
\end{equation}

and

\begin{equation} \label{supp:eq:truestd}
\begin{split}
 \mathbf{\sigma}_{q,c} &= \sqrt{\frac{1}{|D_q|_{c}} \sum_{i,c_i=c}^{N_q} (\mathbf{z_i} - \mathbf{\mu}_{q,c}) \odot (\mathbf{z_i} - \mathbf{\mu}_{q,c}) } \\
 &= \sqrt{\frac{1}{|D_q|_{c}} \sum_{i,c_i=c}^{N_q} (\alpha_{c_i}\zz_i^* + \beta_{c_i} - \alpha_{c_i}\mathbf{\mu}_{q,c}^* - \beta_{c_i}) \odot (\alpha_{c_i}\zz_i^* + \beta_{c_i} - \alpha_{c_i}\mathbf{\mu}_{q,c}^* - \beta_{c_i}) } \\
 &= \sqrt{\frac{\alpha_{c_i}^2}{|D_q|_{c}} \sum_{i,c_i=c}^{N_q} (\zz_i^* - \mathbf{\mu}_{q,c}^*) \odot (\zz_i^* - \mathbf{\mu}_{q,c}^*) } = \alpha_{c_i}\mathbf{\sigma}_{q,c}^*
\end{split}
\end{equation}

This implies that upon normalization, we can write:
\begin{equation} \label{supp:eq:proof}
\begin{split}
 \tilde{\zz}_i &= \frac{\zz_i - \mathbf{\mu}_{q,c}}{\mathbf{\sigma}_{q,c}} \\
 &= \frac{\alpha_{c_i}\zz_i^* + \beta_{c_i} - \alpha_{c_i}\mathbf{\mu}_{q,c}^* - \beta_{c_i}}{\alpha_{c_i}\mathbf{\sigma}_{q,c}^*} \\
 &= \frac{\alpha_{c_i}(\zz_i^* - \mathbf{\mu}_{q,c}^*)}{\alpha_{c_i}\mathbf{\sigma}_{q,c}^*} = \frac{\zz_i^* - \mathbf{\mu}_{q,c}^*}{\mathbf{\sigma}_{q,c}^*} = \tilde{\zz}_i^*
\end{split}
\end{equation}

\autoref{supp:eq:proof} demonstrates that normalization essentially removes the dependence of a bias terms from the effective feature representation completely. Additionally, if we can enforce a criterion that the ideal representation is drawn from a standardized normal distribution i.e. $\tilde\zz^*_i \sim \mathcal{N}(0, 1)$, then the mean and standard deviation can be assumed to be 0 and 1 respectively. This would mean that \autoref{supp:eq:trumean} becomes: $\mu_{q,c}=\beta_{c_i}$, and \autoref{supp:eq:truestd} becomes; $\sigma_{q,c}=\alpha_{c_i}$. Therefore, the mean and standard deviation can be used as a naive estimate for the biasing parameters $\alpha_{c_i}$ and $\beta_{c_i}$. We provide Algorithm \autoref{alg:norm} to improve reproducibility of the method.

We must note that the reliability of the Normalization algorithm rests entirely on the two assumptions: (1) $\tilde\zz^*_i \sim \mathcal{N}(0, 1)$ and (2) $\alpha_{c_i}$ and $\beta_{c_i}$ are constants with respect to camera $c_i$. We argue that for real world data, these assumptions are unrealistic. Assumption (1) may only hold where there are infinitely many person IDs such that the mean of all these features converges at the Origin of the Euclidean space and without a specific standardization, it is not possible to conclude that standard deviation to be unity.

\begin{algorithm}[H]
\caption{Normalization algorithm} \label{alg:norm}
\begin{algorithmic}
\REQUIRE $D_q$, $D_g$, $C$, $f_\theta$
\STATE Initialize $M_q$, $M_g$, $\Sigma_q$, $\Sigma_g$ as empty dictionaries
\STATE Initialize $\Lambda$ as an empty dictionary
\STATE Initialize $Q$, $G$, $C_q$, $C_g$ as empty lists
\FOR{$k=1$ to $C$}
 \STATE $\Lambda(k)=[\text{ }]$
\ENDFOR

\FOR{$i = 1$ to $N_q$}
 \STATE $(\XX_i, p_i, c_i)=D_q(i)$
 \STATE $\zz=f_\theta(\XX_i)$
 \STATE Update $\zz$ to $Q$ and $\Lambda(c_i)$
 \STATE Update $c_i$ to $C_q$
\ENDFOR

\FOR{$i = 1$ to $N_g$}
 \STATE $(\XX_i, p_i, c_i)=D_g(i)$
 \STATE $\zz=f_\theta(\XX_i)$
 \STATE Update $\zz$ to $G$ and $\Lambda(c_i)$
 \STATE Update $c_i$ to $C_g$
\ENDFOR

\STATE Transform lists $Q$, $G$, $C_q$, $C_g$ to tensors

\FOR{$k=1$ to $C$}
 \STATE Transform $\Lambda(k)$ to tensor and overwrite $\Lambda(k)$
 \STATE $M_q(k) = \text{mean}(\Lambda(k))$
 \STATE $M_g(k) = \text{mean}(\Lambda(k))$
 \STATE $\Sigma_q(k) = \text{std}(\Lambda(k))$
 \STATE $\Sigma_g(k) = \text{std}(\Lambda(k))$
\ENDFOR
\STATE $\tilde Q = \frac{Q - M_q(C_q)}{\Sigma_q(C_q)}$
\STATE $\tilde G = \frac{G - M_g(C_g)}{\Sigma_g(C_g)}$
\RETURN $\tilde Q$, $\tilde G$, $M_q$, $M_g$, $\Sigma_q$, $\Sigma_g$
\end{algorithmic}
\end{algorithm}

\begin{figure*}[!h]
 \centering
 \includegraphics[width=0.8\linewidth]{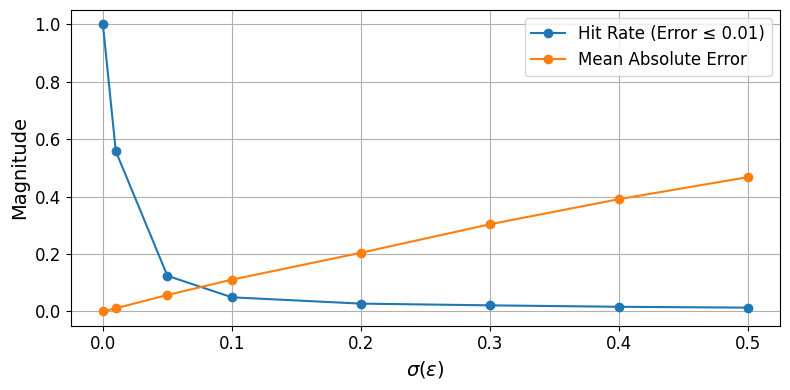}
 \caption{We compare (1) Estimate hit rate (i.e. frequency of correct estimates within 0.01 error threshold) and (2) Mean average error between bias parameter estimates $\alpha_{c_i}$, $\beta_{c_i}$ and true values $\alpha^*_{c_i}$, $\beta^*_{c_i}$. Notice how drastically the hit-rates decrease for even a slight increase in mean average error.}
 \label{supp:fig:error_sim}
\end{figure*}

Assumption (2), on the other hand, can only hold true if and only if the biasing parameters are constant with respect to a camera $c_i$. This assumes that there is no random noise to these biasing parameter estimates. We model a noisy estimate for these parameters, using a noise parameter $\epsilon$, such that $\alpha_{c_i} = \alpha^*_{c_i} + \epsilon_\alpha$ and $\beta_{c_i} = \beta^*_{c_i} + \epsilon_\beta$. The magnitude of deviation $\sigma$ of the noise component $\epsilon = \{\epsilon_\alpha, \epsilon_\beta\} \sim \mathcal{N}(0, \sigma)$ is varied, and we compare the trends of hit rates (i.e. ratio of cases where the estimates are within 0.01 error from true value and total number of test cases) and a mean average error, as shown in \autoref{supp:fig:error_sim}. This simulation demonstrates that even the slightest amount of noise can deviate the estimates of the biasing parameters by a large margin, thereby increasing the prediction uncertainty, thus making camera normalization only a naive solution to address the problem of camera bias.

\subsection{Distance-Aware Retrieval Tuning at Test Time (\method): Algorithm}

\begin{algorithm}[H]
\caption{The \method Algorithm} 
\begin{algorithmic} \label{alg:TTA}
\REQUIRE Image batch $\XX$, Camera IDs $c_x$, Camera statistics $M_q$, $M_g$, $\Sigma_q$, $\Sigma_g$, Source model $f_\theta$, Frozen source model encoded and normalized gallery features $\tilde G$, Number of optimization steps $T$, Euclidean distance computing function $S(\mathbf{A}, \mathbf{B})$, where the shapes of $\mathbf{A} \in \Real^{N_a\times d}$ and $\mathbf{B} \in \Real^{N_b \times d}$.
\ENSURE $M_q$, $\Sigma_q$, and batch-norm parameters of $\theta$ are learnable
\FOR{step=1 to $T$}
 \STATE $\zz = f_\theta(\XX)$
 \STATE $\hat{\zz} = \frac{\zz - M_q(c_x)}{\Sigma_q(c_x)}$
 \STATE $\Delta = \exp{(\frac{-\delta(\hat{\zz}, \tilde{G})}{\tau})}$ // $\tau$ being a temperature parameter
 \STATE $H(\Delta) = -\log(\frac{\Delta}{\sum_{i=1}^{N_g} \Delta(:,i)})$ // softmax operation
 \STATE Top-k distance mask, $M_{top-k}(i,j) = 1$ if $H(\Delta)(i,j)$ is in the top-k least values, else $M_{top-k}(i,j) = 0$
 \STATE $\mathcal{L} = \frac{1}{B} \sum_{i=1}^{B} \sum_{j=1}^{N_g} H(\Delta)(i,j) \cdot M_{top-k}(i,j)$
 \STATE Back-propagate $\mathcal{L}$
 \STATE Update learnable parameters i.e. batch-norm parameters $\mu,\sigma \in \theta$, $M_q$ and $\Sigma_q$.
\ENDFOR
\RETURN $\hat{\zz}$

\end{algorithmic}
\end{algorithm}

\begin{figure*}[!ht]
 \centering
 \includegraphics[width=0.7\linewidth]{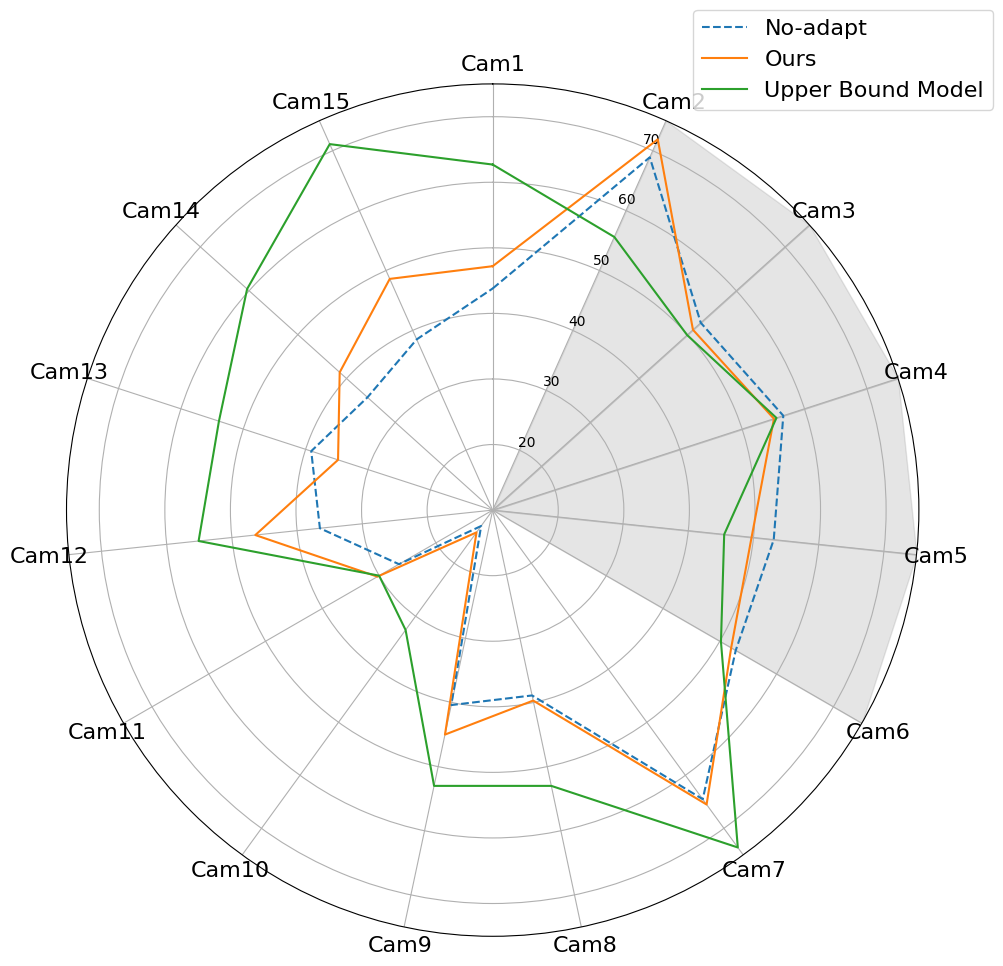}
 \caption{Performance comparison (mAP) for each camera in MSMT17 dataset. Greyed out camera IDs indicate test data from cameras seen by the backbone \cite{li2023clip} during source training. The upper bound model was trained on all cameras but with approximately the same amount of training images as "No-adapt" case. \method consistently performs better on the unseen cameras but the source model, due to its dedicated training on 5 specific cameras, outperforms both the Upper Bound model and \method on the greyed out camera IDs.}
 \label{supp:fig:radial}
\end{figure*}

\section{Additional Experimental Results}

\subsection{Camera ID-wise performance of \method compared to Source model and Upper Bound}

In order to quantify camera specific improvement, we tabulate the evaluation metrics for \method, compared to the original source model and the upper bound in \autoref{supp:fig:radial}. The source model is trained on data from 5 cameras, in the MSMT17 dataset and is evaluated on all the cameras separately. Specifically, we use cameras 2 to 6 for the source pretraining. The upper bound model uses data from all 15 cameras, but the total number of training images is reduced to match that of the number of images from the 5 cameras (i.e. IDs 2, 3, 4, 5 and 6). 

A striking observation is that the source model outperforms our assumed upper bound model in the cameras that have been used for source pretraining, despite the exact images and person identities being different in the test set. This can be explained by the fact that the source model is dedicatedly trained on only 5 cameras, as a result of which it performs well on these samples. The upper bound model on the other hand consistently performs well on all the 15 camera IDs. Except for Camera ID 13, our method outperforms the source model on all other unseen cameras.

\subsection{Effect of Hyperparameters for \method and \method LITE}

\begin{figure}[!ht]
    \centering
    \begin{subfigure}[t]{0.31\textwidth}
        \includegraphics[width=\linewidth]{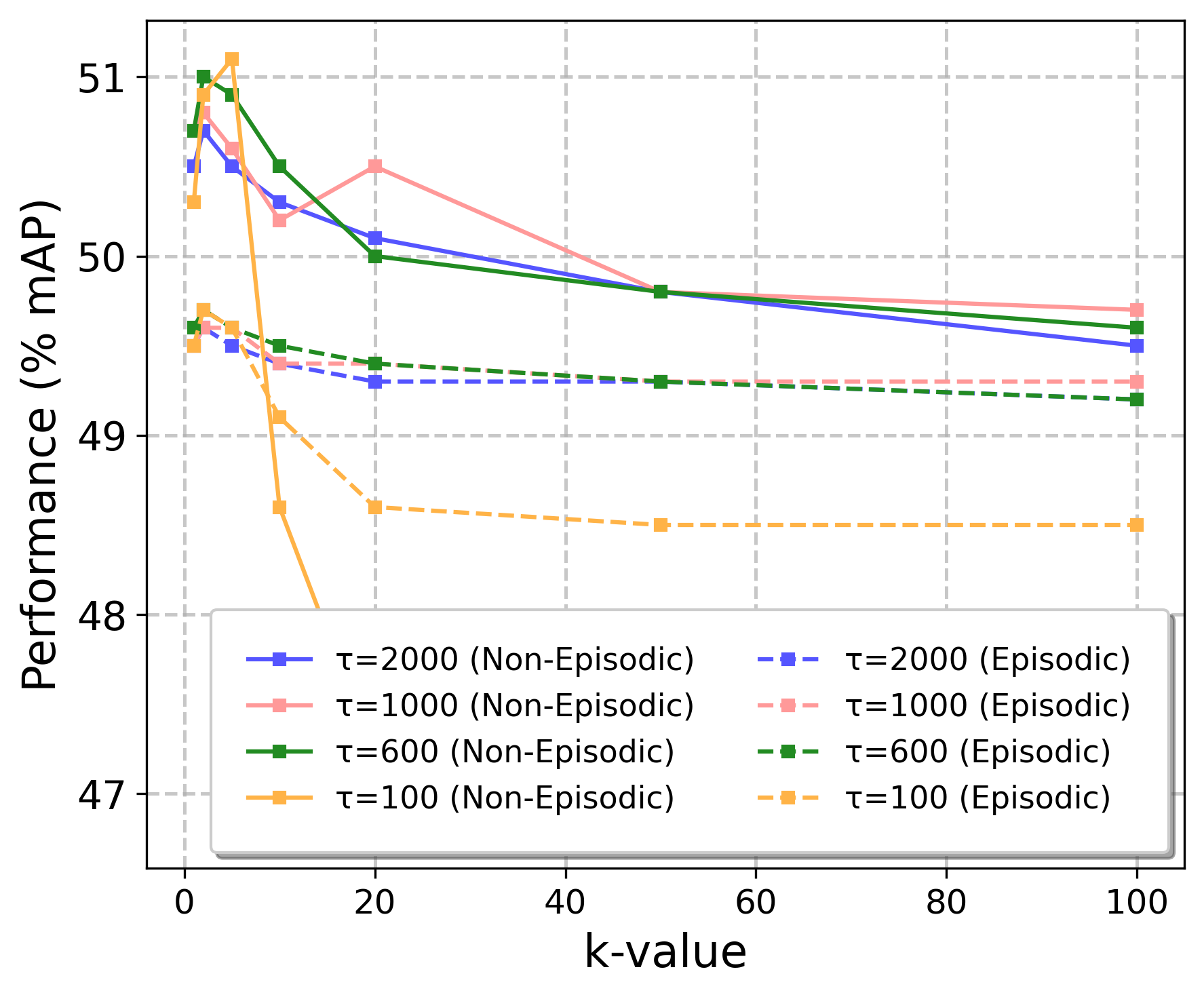}
        \caption{} \label{supp:fig:abl_k}
    \end{subfigure}
    \hfill
    \begin{subfigure}[t]{0.31\textwidth}
        \includegraphics[width=\linewidth]{figures/abl_temp.png}
        \caption{} \label{supp:fig:abl_tau}
    \end{subfigure}
    \hfill
    \begin{subfigure}[t]{0.31\textwidth}
        \includegraphics[width=\linewidth]{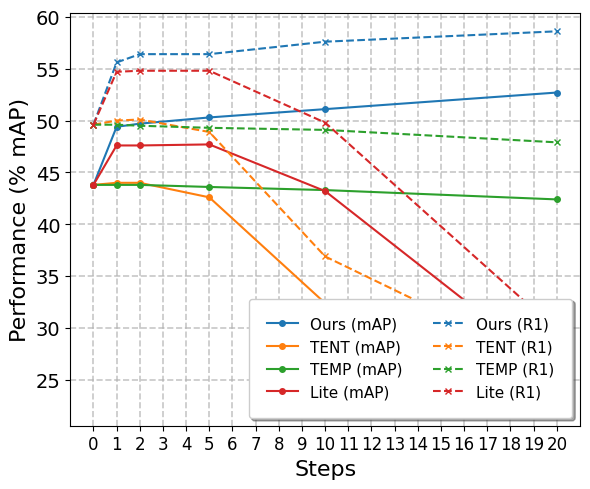}
        \caption{} \label{supp:fig:abl_ne_step}
    \end{subfigure}
    
    \vspace{1em} 

    \begin{subfigure}[t]{0.31\textwidth}
        \includegraphics[width=\linewidth]{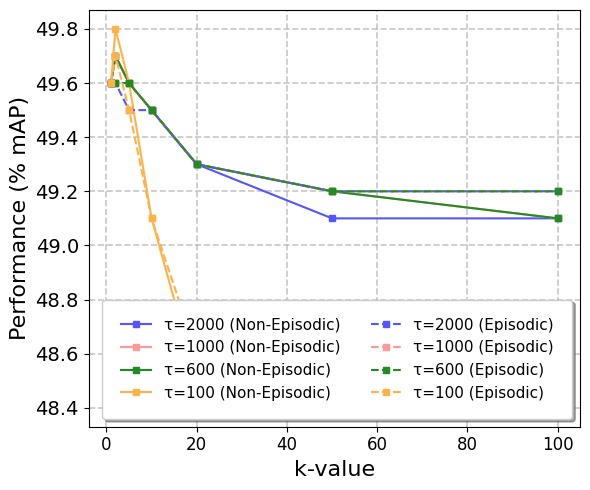}
        \caption{} \label{supp:fig:abl_k_lite}
    \end{subfigure}
    \hfill
    \begin{subfigure}[t]{0.31\textwidth}
        \includegraphics[width=\linewidth]{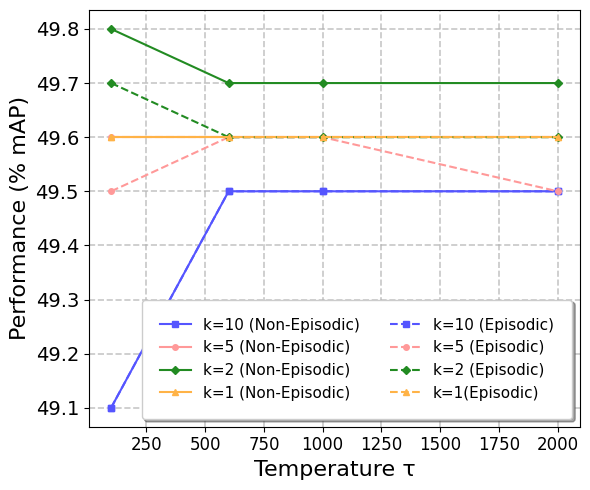}
        \caption{} \label{supp:fig:abl_tau_lite}
    \end{subfigure}
    \hfill
    \begin{subfigure}[t]{0.31\textwidth}
        \includegraphics[width=\linewidth]{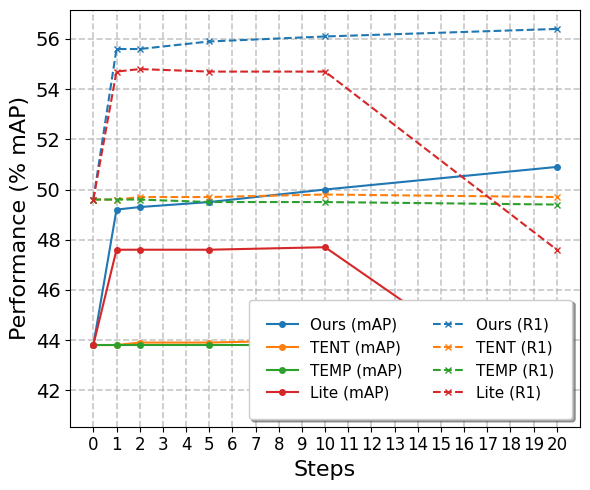}
        \caption{} \label{supp:fig:abl_e_step}
    \end{subfigure}

    \caption{Trends in performance (mAP) with respect to the value of (a) $k$, (b) $\tau$ and number of steps of optimization per batch for (c) non-episodic training and (f) episodic training. (d) and (e) shows the trends in performance for varying $k$ and $\tau$ values for \method LITE.}
    \label{fig:multi_panel}
\end{figure}

\mypar{Effect of $k$: } The selection of top-k Euclidean distances within a batch, is intended to soften the penalization of the loss over the batch of queries. For both episodic and non-episodic training, we find that a $k$ value between 1 to 5 performs the best, as can be seen in \autoref{supp:fig:abl_k}. However, choosing a larger $k$ value shows that the scores increase with increasing temperature value. In general, it appears to be a good rule of thumb to choose a smaller $k$ value if a lower temperature is chosen, and vice-versa. We believe that this effect is seen largely due to lower temperature scores making the differences in distances larger, and as a result, it helps to have a smaller $k$ value, since it would prevent over-penalization. In contrast, a higher temperature leads to lower distances and a small $k$ value will lead to the loss magnitude being insufficiently large to adapt. In the case of \method LITE, in \autoref{supp:fig:abl_k_lite} we observe that the range of performance variation is smaller than \method, but a top k value of 2-5 is still consistently better. However, the trends in performance are roughly similar to that of \method.

\mypar{Effect of Softmax Temperature $\tau$:} Unlike classification confidence scores, the ReID model predicts spatial coordinates which are unbounded in nature. As a result, the temperature values used in our task is unusually large, in a range between 100-2000. For small temperature values, the resultant exponential value diminishes below the precision limit of the hardware used, thereby resulting in a collapse. This hyperparameter seems to only soften the effects of adaptation in multiple steps. From \autoref{supp:fig:abl_tau}, we observe that a higher temperature score leads to a more stable trend in variations of performance over multiple steps and various values of $k$, but leads to a lower peak than to a lower value. Since its not very practical to train a TTA pipeline for more than 1-5 steps per batch, we choose a temperature of 100-200 for most of our experiments. However, in the case of \method LITE \autoref{supp:fig:abl_tau_lite} shows that the variation at higher temperature values become roughly constant. In general, the best values for temperature for \method LITE are between 100-200 for k values less than 5.

\mypar{Effect of Number of Steps: }
The number of steps is essentially the number of steps a batch of queries is optimized for. We find that the performance of our method consistently improves with the number of steps both in the episodic and non-episodic settings. \autoref{supp:fig:abl_ne_step} and \autoref{supp:fig:abl_e_step} demonstrates the two cases. We find that in comparison, entropy based methods TENT \cite{wang2020tent} and TEMP \cite{adachi2024test} deteriorate drastically over increasing number of steps in the non-episodic setting. \method LITE, however, because it is a case where only a few parameters train, over-training can lead to serious overfitting, deteriorating the effects of source pretraining. This effect is seen by performances deteriorating in a striking manner beyond 5-10 steps per batch.

\begin{figure*}[!ht]
 \centering
 \includegraphics[width=0.8\linewidth]{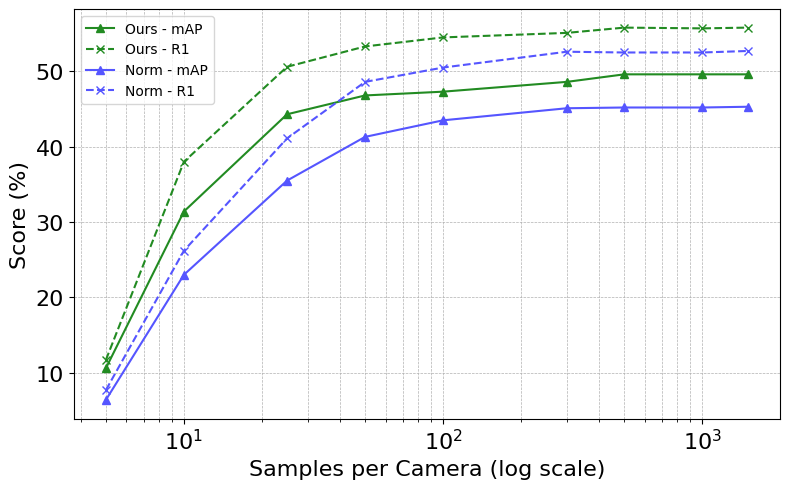}
 \caption{Comparison of performance with respect to number of samples used per camera to estimate normalization statistics. \method consistently performs better even at lower number of samples per camera, due to its adaptive nature.}
 \label{supp:fig:samples_per_cam}
\end{figure*}

\subsection{Effect of Number of samples per camera in Normalization Dictionaries $M$ and $\Sigma$}

We explore the impact of the number of sample images used for computation of the initial normalization statistics. For this experiment, we sample an equal number of images per camera sample to compute the mean and standard deviation, for MSMT17 dataset for a single unseen camera. Our source model is a pretrained CLIP-ReID backbone \cite{li2023clip}, trained on only 5 cameras from MSMT17 dataset. \autoref{supp:fig:samples_per_cam} shows our findings. We observe that \method achieves a more consistent improvement over Camera Normalization \cite{song2025exploring}, due to its adaptive nature. In summary, \method requires fewer images per camera to compute its initialization parameters.

\end{document}